\documentclass[11pt]{article}

\usepackage{array}
\usepackage{amsmath}
\newcolumntype{P}[1]{>{\centering\arraybackslash}p{#1}}
\newcolumntype{M}[1]{>{\centering\arraybackslash}m{#1}}

\usepackage{graphicx}
\graphicspath{ {figure/} }
\usepackage{float}
\usepackage{subcaption}
\usepackage{listings}\usepackage[shortlabels]{enumitem}
\usepackage{comment}

\usepackage{array}
\usepackage{multirow}
\usepackage{tabularx}

\usepackage{hyperref}

\usepackage{amsthm}
\theoremstyle{definition}
\newtheorem{definition}{Definition}[section]
\usepackage{algorithm2e}
\usepackage{authblk}

\author[1]{Md Imran Hossen}

\author[2]{Sai Venkatesh Chilukoti} 

\author[3]{Liqun Shan}

\author[4]{Vijay Srinivas Tida}

\author[5]{Xiali Hei}

\affil[1]{University of Louisiana at Lafayette, Lafayette, Louisiana, USA, md-imran.hossen1@louisiana.edu}
\affil[2]{University of Louisiana at Lafayette, Lafayette, Louisiana, USA, sai-venkatesh.chilukoti1@louisiana.edu}
\affil[3]{University of Louisiana at Lafayette, Lafayette, Louisiana, USA, liqun.shan@louisiana.edu}
\affil[4]{College of St. Benedict and St. John's University, St. Joseph, Minnesota, USA, vtida001@csbsju.edu}
\affil[5]{University of Louisiana at Lafayette, Lafayette, Louisiana, USA, xiali.hei@louisiana.edu}

\begin{document}

% ========== Edit your name here
%\author{PI: Dr. Xiali Hei}

\title{Facebook Report on Privacy of fNIRS data}
\maketitle

\medskip
\section{Project Goal}

\paragraph{Objectives.} 
The primary goal of this project is to develop privacy-preserving machine learning model training techniques for fNIRS data.
This project will build a local model in a centralized setting with both differential privacy (DP) and certified robustness. It will also explore collaborative federated learning to train a shared model between multiple clients without sharing local fNIRS datasets. To prevent unintentional private information leakage of such clients' private datasets, we will also implement DP in the federated learning setting.
%without retraining the pre-trained learning model. 
%We assume that the model has finished pre-training from public fNIRS datasets.
% Our goal is to train a noise remover based on the user's fNIRS dataset
% and add it to the pre-trained model to achieve both differential privacy and certified robustness of the final model. 

\paragraph{Motivation.} 
% \textbf{Background}:
VR/AR devices usually employ multi-modality sensors (fNIRS, motion, etc.) to achieve spatial awareness. All sensed, associated, and computed data contain rich patterns. However, adversaries might obtain the private and sensitive information of users by using some special devices / methods for data access \cite{dick2021balancing}. For example, eye tracking has been used to help mental health practitioners diagnose certain brain disorders \cite{disorder}.
Today's AR/VR devices can access the eye tracker integrated with the brain control interface. From the collected fNIRS data, an adversary can obtain the user's private data, including heart rate variability \cite{hakimi2018stress, hakimi2020proposing, perdue2014extraction}, attention information \cite{jahani2015attention, durantin2015characterization, zhang2016passive, murata2015culturally}, Alzheimer's disease information \cite{li2018early, li2019dynamic, cicalese2020eeg}, and epileptic seizure events \cite{rosas2019prediction, irani2007functional, rizki2015determination}. 

Due to the sensitive nature of fNIRS data, direct sharing options for such datasets are severely limited. To aid research, one might release a deep neural network (DNN) model that has been pre-trained on a private dataset. However, deep neural networks include a large number of hidden layers, resulting in a high effective capacity that could be enough to encode the details of some individual's data into model parameters or even memorize the entire dataset \cite{zhang2021understanding, song2017machine}. Furthermore, when model parameters are exposed, adversaries can use membership attacks \cite{truex2019demystifying, nasr2018comprehensive, shokri2017membership} or model inversion attacks \cite{fredrikson2015model} to infer sensitive data records of individuals in private training datasets. Even if only query APIs are provided to access remote trained models, model parameters can be retrieved from prediction queries and utilized to infer sensitive training data \cite{tramer2016stealing}. As a result, it is critical to develop fundamental privacy-preserving deep learning approaches to protect private fNIRS training data from attackers with partial or complete knowledge of model parameters.

\textbf{New approaches to privacy-preserving machine learning for fNIRS data.}
In centralized deep learning, membership inference \cite{truex2019demystifying, nasr2018comprehensive, shokri2017membership} can reveal private training data by leveraging statistical differences between model predictions on data possibly seen during training and predictions on unseen data.
To defend against such attacks and prevent malicious parties, such as dishonest servers, from inferring sensitive information (\textit{e.g.} heartbeat rate, seizure, etc.) from a model trained on fNIRS data, we will add noise (also called perturbation) to achieve differential privacy (DP) \cite{dwork2011firm, dwork2006calibrating, dwork2014algorithmic}. Differential privacy (DP) is a provable and quantifiable method of privacy protection that makes it virtually impossible for an adversary to distinguish between the results obtained from two neighboring datasets. Neighboring/adjacent datasets are datasets that differ only by a single data sample. 

% There are four types of perturbations: input perturbation, output perturbation, gradient perturbation, objective perturbation, and noisy labeling. Input perturbation is easy for differential privacy implementation, but it is challenging to achieve high model accuracy due to the excessive noise injection.
%Gradient perturbation is one of the most popular approaches to achieve differential privacy in deep learning models. 
% Randomized smoothing \sqcitet{lee2019tight, li2018certified} is one of the most popular techniques to achieve certified robustness by adding random perturbations to the inputs and then calculating the
% robustness bound.

We will also investigate the effectiveness of the federated learning approach for training models on fNIRS datasets. FL moves the learning task away from the centralized server and onto a distributed network of client nodes \cite{mcmahan2017communication,vanhaesebrouck2017decentralized, konevcny2016federated, bonawitz2019towards}. FL enables clients to collaboratively learn a shared prediction model without sharing data. As such, it solves some of the critical data security and data privacy issues found in centralized machine learning.

% It learns a shared global model collaboratively from a distributed collection of training data stored on the end-user devices of the participating clients. At each round of federated learning, each client downloads the global model update from the server, performs local training, and shares its local model update with the server via encrypted communication.

However, the distributed nature of FL makes it vulnerable to gradient leakage attacks, so the default privacy protection in FL is insufficient for protecting client training data. 
Recent research \cite{zhu2019deep, geiping2020inverting} reveals that if an adversary intercepts a client's local gradient update during a global round of federated learning, the adversary can steal the sensitive local training data of this client using the leaked gradients by simply performing a reconstruction attack \cite{wei2020framework}, or membership and attribute inference attacks \cite{truex2019demystifying, nasr2018comprehensive, shokri2017membership}. To prevent such attacks, we will incorporate differential privacy (DP) into the FL setting. Specifically, we will apply DP to prevent third parties and honest but curious global servers from inferring clients' private training data from the local model updates shared by the clients. 

% However, the distributed nature and the default privacy in federated learning are insufficient for protecting client training data from gradient leakage attacks. Recent studies [6]–[8] reveal that if an adversary intercepts the local gradient update of a client before the server performs the federated aggregation to generate the global parameter update for the next round of federated learning, the adversary can steal the sensitive local training data of this client using the leaked gradients by simply performing reconstruction attack [8], or membership and attribute inference attacks [9]–[11].

\section{Background}

\paragraph{Differential Privacy (DP)}
% Differential privacy (DP) provides a well-tested formalization for the release of information derived from private data. Applied to machine learning, a differentially private training mechanism allows the public release of model parameters with a strong guarantee: adversaries are severely limited in what they can learn about the original training data based on analyzing the parameters, even when they have access to arbitrary side information.  Formally, it says:
Differential privacy (DP) is a formal mathematical framework for defining the privacy properties of data analysis algorithms. 
% Informally, it states that any changes to a single data point in the training dataset should only result in statistically insignificant changes to the algorithm's output.
When used in machine learning, a differentially private training mechanism enables the release of model parameters with a strong privacy guarantee: adversaries are severely constrained in what they can learn about the original training data by examining the parameters, even if they have access to arbitrary side information. Formally, it is defined as follows:

\begin{definition}[Differential Privacy]
A randomized mechanism $\mathcal{M}: \mathcal{D} \rightarrow \mathcal{R}$  with a domain $\mathcal{D}$ (\textit{e.g.},
possible training datasets) and range $\mathcal{R}$ (\textit{e.g.}, all possible trained models) satisfies ($\epsilon$, $\delta$)-differential
privacy if for any two datasets $d, d\;' \in \mathcal{D}$, differing with only a single data sample and for any subset of outputs $\mathcal{S} \subseteq \mathcal{R}$ it holds that
$Pr[\mathcal{M}(d) \in \mathcal{S}] \leq e^\epsilon Pr[\mathcal{M}(d\;') \in \mathcal{S}] + \delta$.
\end{definition}

%  According to most prior work on differentially private machine learning two datasets $d$ and $d\;'$ are defined to be \textit{adjacent} if $d\;'$ can be formed by adding or removing a single training example from $d$. If $\delta=0$, $\mathcal{M}$ is said to be $\epsilon$-differential privacy, which is stronger but less flexible in terms of the mechanism design.

\paragraph{Deep learning with differential privacy (DP)}
% DP characterizes the difference in output between two input datasets differing by
% at most one element.
% Because it is difficult to characterize the maximum difference of the model parameters over any two neighboring datasets for neural networks, differentially private
% deep learning \cite{mcmahan2017learning, abadi2016deep, song2013stochastic} relies on differentially private stochastic gradient descent (DP-SGD) to control the influence of training data on the model.

In 2016, Abadi \textit{et al}. \cite{abadi2016deep} presented the first proposal for deep learning with differential privacy. Differentially private deep learning \cite{mcmahan2017learning, abadi2016deep, song2013stochastic} typically relies on differentially private stochastic gradient descent (DP-SGD) to control the influence of training data on the model.

% DP-SGD explicitly bounds per-example gradients $\Delta_{\theta}L(\theta, x)$ in every iteration by clipping the $L_2$ norm of gradient vectors. Given a clipping threshold S, this is done by replacing the gradient vector $\textbf{g}$ with $\textbf{g}/$max(1, $\frac{||\textbf{g}||_2}{S}$) which scales \textbf{g} down to norm $S$ if $||\textbf{g}||_2 > C$. A Gaussian mechanism with $L_2$ norm sensitivity of $S$ is then applied to perturb the gradients before the gradient descent step updates the model parameter.

DP-SGD \cite{abadi2016deep} works as follows: At each step of the SGD, it computes the gradient $\Delta_{\theta}L(\theta, x_i)$ for a random subset of examples, computes the per-sample gradient, clips the $L_2$ norm of each gradient, computes the average, adds noise to protect privacy, and takes a step in the opposite direction of this average noisy gradient. The whole process is illustrated in Algorithm \ref{alg:dpsgd}. Gradient clipping is required when implementing DP-SGD because the amount of information extracted from a dataset for an individual sample is proportional to the magnitude of the gradient. Therefore, if gradient clipping is not applied, the amount of added noise should be larger to protect the privacy of individual samples, which degrades the utility of the model.

\RestyleAlgo{ruled}
\SetKwComment{Comment}{/* }{ */}
\begin{algorithm}[!tp]
% local change
\SetKwInput{KwData}{Input}
\SetKwInput{KwResult}{Output}
\caption{Differentially private SGD (DP-SGD)}\label{alg:dpsgd}
\KwData{Examples \{$x_1, ...,x_N$\}, loss function $\mathcal{L}(\theta) = \frac{1}{N}\sum_i\mathcal{L}(\theta, x_i)$. Parameters: learning rate $\eta_t$, noise scale
$\sigma$, group size $L$, gradient norm bound $S$.}
\textbf{Initialize} $\theta_0$ randomly\\
\For{$1 \in [T]$}{
    Take a random sample $L_t$ with sampling probability $L/N$\\
    \textbf{Compute gradient}\\
    For each $i \in L_t$, compute $\textbf{g}_t(x_i) \gets \Delta_{\theta_{t}}\mathcal{L}(\theta_t, x_i)$\\
   \textbf{Clip gradient}\\
    $\bar{\textbf{g}}_t(x_i)$ $\gets$ $\textbf{g}_t(x_i)$/max(1, $\frac{||\textbf{g}_t(x_i)||_2}{S})$\\
    \textbf{Add noise}\\
    $\tilde{\textbf{g}}_t \gets \frac{1}{L}(\sum_i(\bar{\textbf{g}}_t(x_i) + \mathcal{N}(0, \sigma^2S^2\mathbf{I}))$\\
   \textbf{Descent}\\
    $\theta_{t+1} \gets \theta_t - \eta_{t}\tilde{\textbf{g}}_t$\\
}
\KwResult{$\theta_T$ and compute the overall privacy $(\epsilon, \delta)$ 
using a privacy accounting method.}
\end{algorithm}

\subsection{Federated Learning (FL)} Federated learning (FL) involves training a machine learning algorithm, such as deep neural networks, on multiple local datasets stored in local client nodes without explicitly exchanging data samples. Every client has a locally stored training dataset that is never uploaded to the server, which is responsible for orchestrating the training. Instead, each client computes an update to the server's current global model, and only that update (for example, a deep neural network's weights and biases) is communicated.
A fundamental advantage of FL is that it decouples model training from the necessity for direct access to the raw training data, allowing for critical issues like data privacy, data security, and data access rights to be addressed.

\paragraph{Federated aggregation}
The typical federated learning paradigm consists of two stages: clients train the models separately using their local datasets, and the data center aggregates the locally trained models to produce a shared global model. A common aggregation technique called federated averaging (FedAvg) \cite{mcmahan2017communication} averages the local model parameters element-wise with weights proportional to the sizes of the client datasets. Algorithm \ref{alg:fedavg} shows how FedAvg works. Another more common aggregation algorithm is FedProx, \cite{sahu2018convergence}  %reduces the impact of local updates by keeping them close to the global model by incorporating a proximal term into the client cost functions. FedProx 
which is shown to tackle both the system and statistical heterogenity theoretically as well as empirically~\cite{li2020federated}. FedProx can be viewed as a generalization of FedAvg. FedProx makes a simple modification to the FedAvg algorithm that is an addition of the proximal term to the local client's optimization task. The local solver objective function can be explained using Equation \ref{eq04}:

\begin{equation}\label{eq04}
\begin{array}{l}
    \min_{w}h_k(w; w^t)= F_k(w) + \frac{\mu}{2}||w - w^t||^2.
\end{array}    
\end{equation}

 Where $w^t$ represents the weights of the initial global model, $w$ represents the local client model weights, $\mu$ is a hyperparameter which controls the amount of proximal term added to the client $k$'s local objective function $F_k(w)$. The proximal term has two advantages: (1) it ensures that the local updates are as close as possible to the initial global model to tackle the statistical heterogeneity; and (2) it also allows local clients to perform variable amounts of work resulting from system heterogeneity.

\RestyleAlgo{ruled}
%% This is needed if you want to add comments in
%% your algorithm with \Comment
\SetKwComment{Comment}{/* }{ */}
{\SetAlgoNoLine%
\begin{algorithm}[tp!]
% local change
\SetArgSty{textnormal}
\SetKwInput{KwData}{Input}
\SetKwInput{KwResult}{Output}
\caption{Federated averaging (FedAvg) algorithm. The $K$ clients are indexed by $k$; $B$ is the local minibatch size, $E$ is the number of local epochs, and $\eta$ is the learning rate.}\label{alg:fedavg}
% \KwData{The $K$ clients are indexed by $k$; $B$ is the local minibatch size, $E$ is the number of local epochs, and $\eta$ is the learning rate.}
 \textbf{Server executes:}\\
\Indp initialize $w_0$\\
\For{each round $t = 1, 2, . . .$}{
{$m \gets$ max$(C\cdot K, 1)$}\\
$S_t \gets$ (random set of $m$ clients)\\
\For{each client $k \in S_t$ in parallel}{
$w_{t+1}^k \gets$ ClientUpdate($k, w_t$)\\
}
$w_{t+1} \gets \sum_{k=1}^K \frac{n_k}{n}w_{t+1}^k$\\
}
~\\
\Indm \textbf{ClientUpdate}($k, w$)\textbf{:} // Run on client $k$\\
\Indp
$\mathcal{B}  \gets$ (split $\mathcal{P}_k$ into batches of size $\mathcal{B}$)\\
\For{\text{each local epoch} $i$ from $1$ to $E$}{
\For{batch $b \in \mathcal{B}$}{
  $w \gets w - \eta\Delta \ell(w; b)$\\
  }
}
return $w$ to server\\
% \KwResult{$\theta_T$ and compute the overall privacy $(\epsilon, \delta)$ 
% using a privacy accounting method.}
\end{algorithm}
}

\section{Experimental results}

\paragraph{Dataset} We used the Tufts fNIRS to Mental Workload (fNIRS2MW) data set \cite{huangfNIRS2MW2021}, which contains multivariate fNIRS recordings from 68 participants, each with labeled segments indicating four possible levels of intensity of mental workload. We use the data set for binary classification purposes, and each data sample has one of two associated levels of intensity of mental workload (the lowest and the highest) as described in the paper. To keep the evaluation simple, we split the data set into train, validation, and test sets with a ratio of 0.6, 0.2, and 0.2, respectively.

\paragraph{Implementation and evaluation platform}
All of our experiments are carried out on a machine with 36 Intel Core i9-10980XE CPUs, 251 GB of memory, and an NVIDIA Quadro RTX 8000 GPU running the Ubuntu 18.04 LTS operating system. The code was developed using PyTorch. We use the Opacus library \cite{yousefpour2021opacus} to implement and train models with differential privacy (DP).

% \begin{table}[tp!]
%     \centering
%     \caption{Performance of different models.}
%     \label{tab:my_label}
%     \begin{tabular}{|c|c|c|c|c|}
%         \hline
%         Model & Accuracy & Precision & Recall & F1-score  \\
%         \hline
%         MLP &  0.70 & 0.70 & 0.70 & 0.70 \\
%         DeepConvNet & 0.65 & 0.66 &  0.65 & 0.64  \\
%         EEGNet & 0.67 & 0.67 & 0.67 & 0.66 \\
%         LSTM & 0.66 & 0.66 & 0.66 & 0.66 \\
%         BiLSTM & 0.66 & 0.66 &  0.66 & 0.66\\
%         \hline
%     \end{tabular}
% \end{table}

\subsection{Non-private centralized training (baseline)}
As a baseline, we train several deep learning models in a non-private fashion for the metal workload classification task on the Tufs fNIRS mental workload dataset. We opted to use only deep neural network models for classification since they do not involve a laborious hand-crafted feature engineering process for training, unlike traditional machine learning models. In particular, we tested the following models: 1) a Multilayer Perceptron (MLP) containing the input layer, three hidden layers, and the output layer; 2) DeepConvNet \cite{schirrmeister2017deep}, 3) a simple LSTM network, and 4) a BiLSTM network.

\begin{figure}[htb]
\centering
%\captionsetup[subfigure]{font=small} if you like to change caption style
     \begin{subfigure}[b]{0.45\textwidth}
         \includegraphics[width=\textwidth]{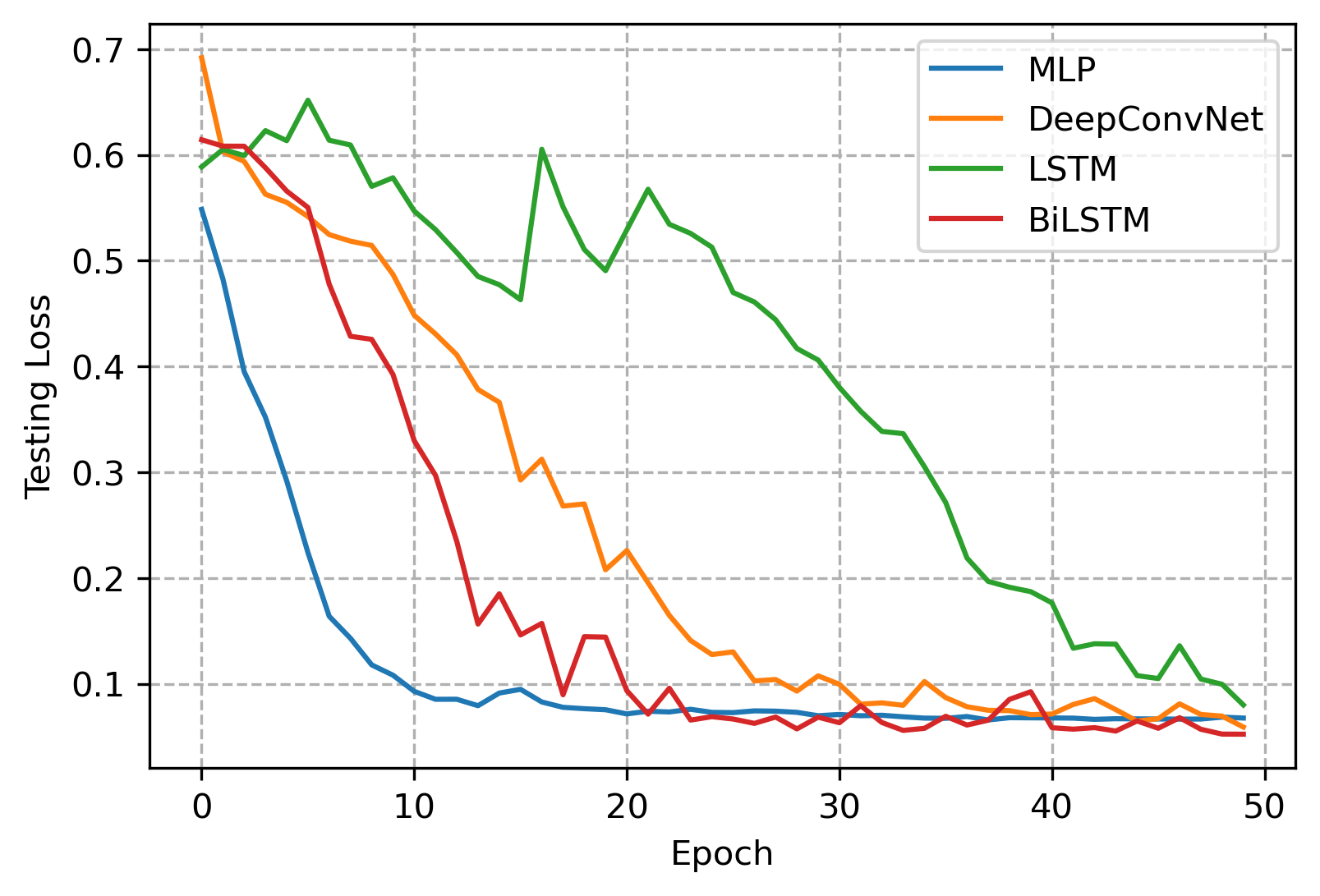}
         \caption[]{Training loss} % <---
         \label{fig:CT_no_dp:train_loss}
     \end{subfigure}
    %  \hfill
     \begin{subfigure}[b]{0.45\textwidth}
         \includegraphics[width=\textwidth]{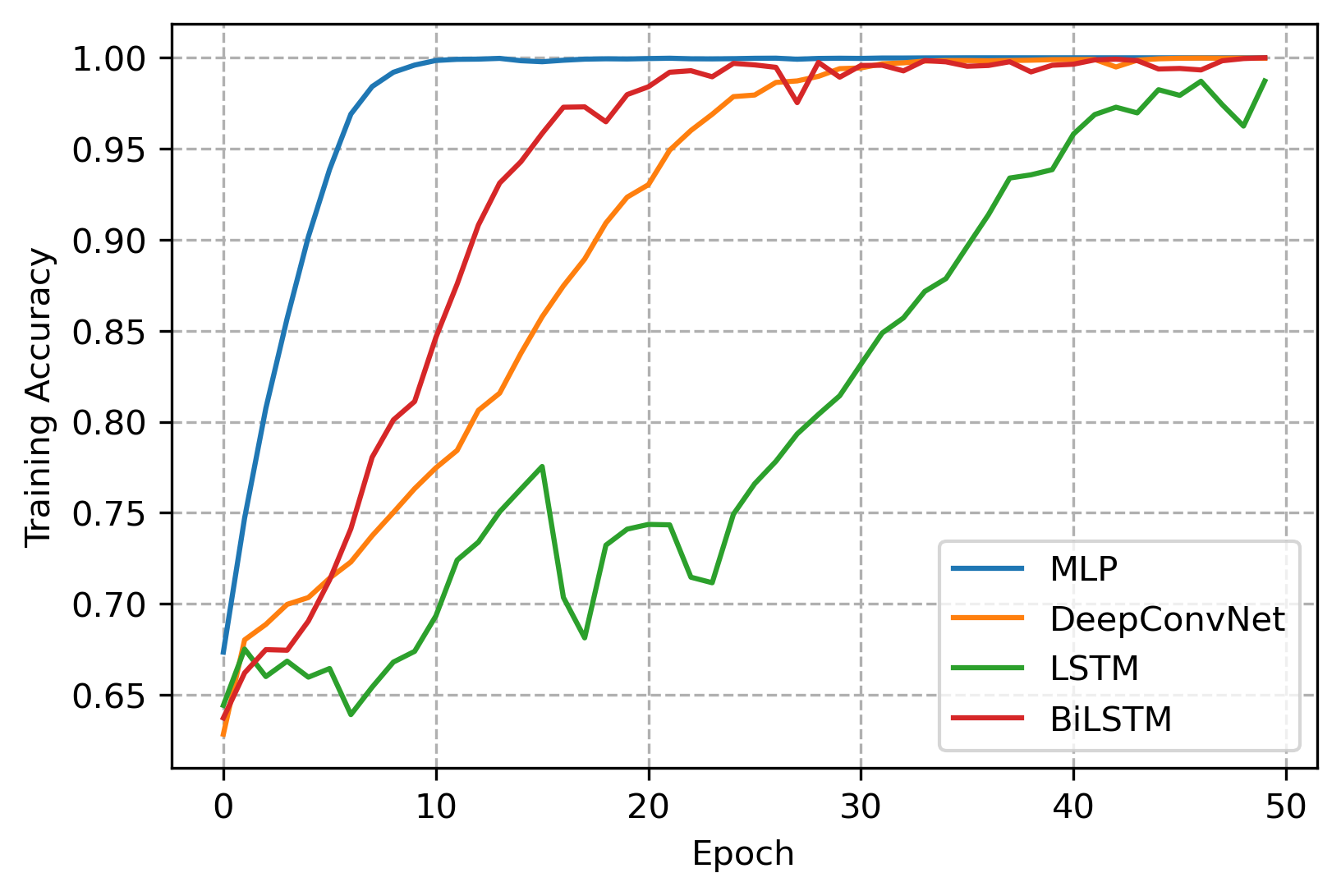}
         \caption[]{Training accuracy} % <---
         \label{fig:CT_no_dp:train_acc}
     \end{subfigure}

    %  \vskip\baselineskip
     \begin{subfigure}[b]{0.45\textwidth}
         \includegraphics[width=\textwidth]{CT_noDP_test_loss.png}
         \caption[]{Testing loss} % <---
         \label{fig:CT_no_dp:test_loss}
     \end{subfigure}
    %  \hfill
     \begin{subfigure}[b]{0.45\textwidth}
         \centering
         \includegraphics[width=\textwidth]{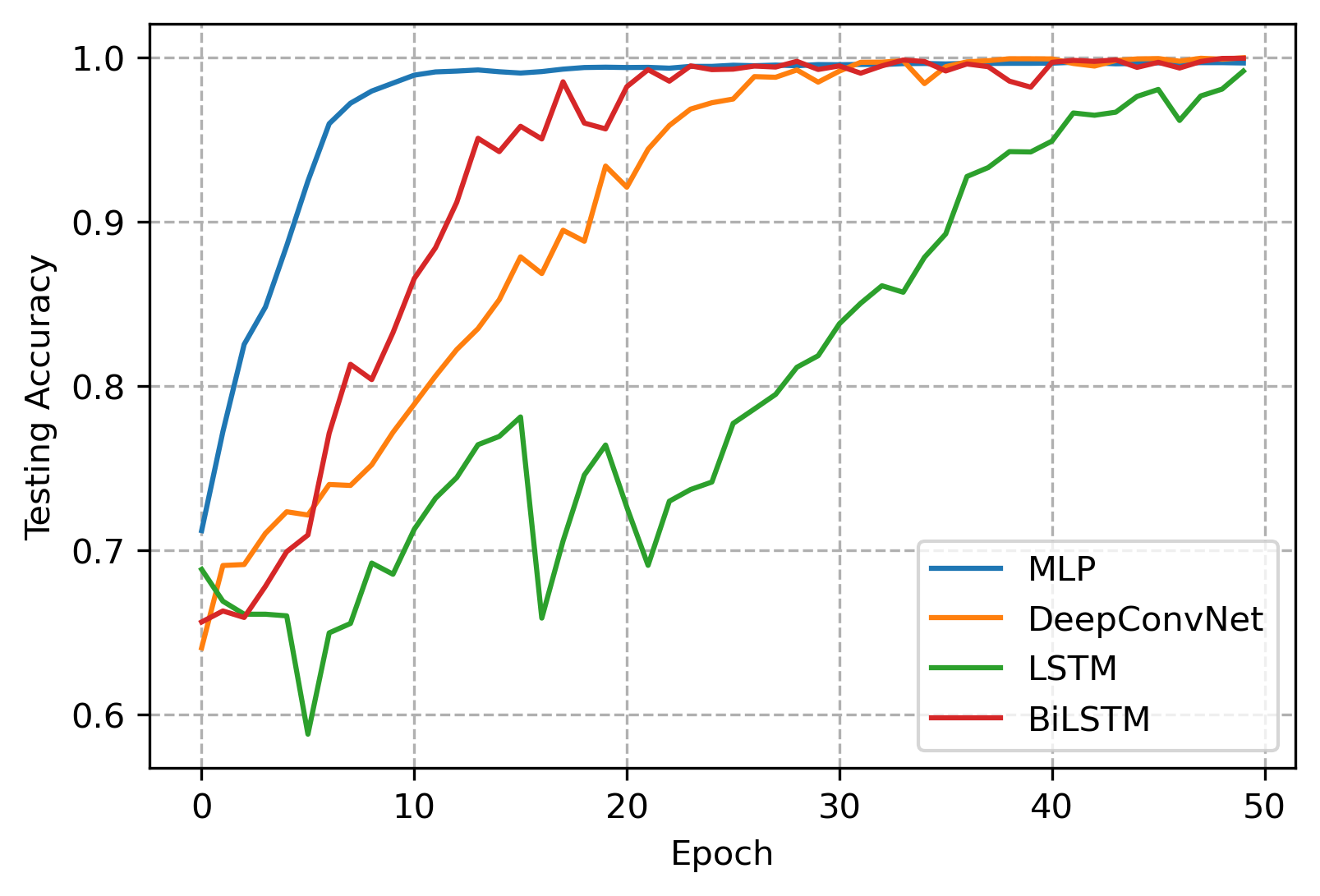}
         \caption[]{Testing accuracy} % <---
         \label{fig:CT_no_dp:test_acc}
     \end{subfigure}
        \caption{Training curves for different models.} % <---
        \label{fig:CT_no_dp:model_perf}
\end{figure}

We conduct hyperparameter searches for each model separately. We find that the Adam optimizer performs better than SGD in most cases, so all models are trained using the Adam optimizer. We use a batch size of 1024 for the MLP models and 256 for the remaining models. The learning rate of 0.001 appears to be optimal for the MLP, LSTM, and BiLSTM models. For DeepConvNet, we used a learning rate of 0.01. We train the models for 50 epochs. Figure \ref{fig:CT_no_dp:model_perf} shows the training curves for the models. All models obtain a test accuracy of more than 99\%. However, the simple MLP converges faster and its training and testing losses decrease more consistently compared to other models.

In the rest of the paper, we will use the MLP network for all of the experiments.

\subsection{Differentially private centralized training}

For the differentially private version, we experiment with the same MLP model. The accuracy of classification is influenced by a number of factors, all of which need to be carefully tuned for optimum performance. The network architecture, the DP optimization algorithm, as well as parameters of the training procedure like the minibatch size and learning rate are some of these factors. Some parameters are specific to privacy, such as clipping bound $S$ and noise multiplier $\sigma$.

To demonstrate the effects of these parameters, we modify each one individually while keeping the rest constant. We set the reference values as follows: batch size of 2048, gradient norm bound $S$ of 4, learning rate of 0.003, epochs of 50, and noise multiplier $\sigma$ equal to 0.8. For each combination of values, we train up to the point where (22.59, $10^{-5}$)-differential privacy would be violated. So, for instance, a larger $\sigma$ allows more epochs of training.

\paragraph{Differentially private optimization algorithm} Since optimization using Adam performs better than SGD for non-private training for the dataset we used in this paper, we investigate the training performance of differentially private versions of both of these optimizers for our private training. We perform a parameter search for DP-SGD and DP-Adam separately. We train for 50 epochs to reach (22.59, $10^{-5}$)-DP using a batch size of 2048 for both and learning rates of 0.05 and 0.003 for DP-SGD and DP-Adam, respectively.

\begin{figure*}[!tp]
\centering
%\captionsetup[subfigure]{font=small} 
     \begin{subfigure}[b]{0.32\textwidth}
         \includegraphics[width=\textwidth]{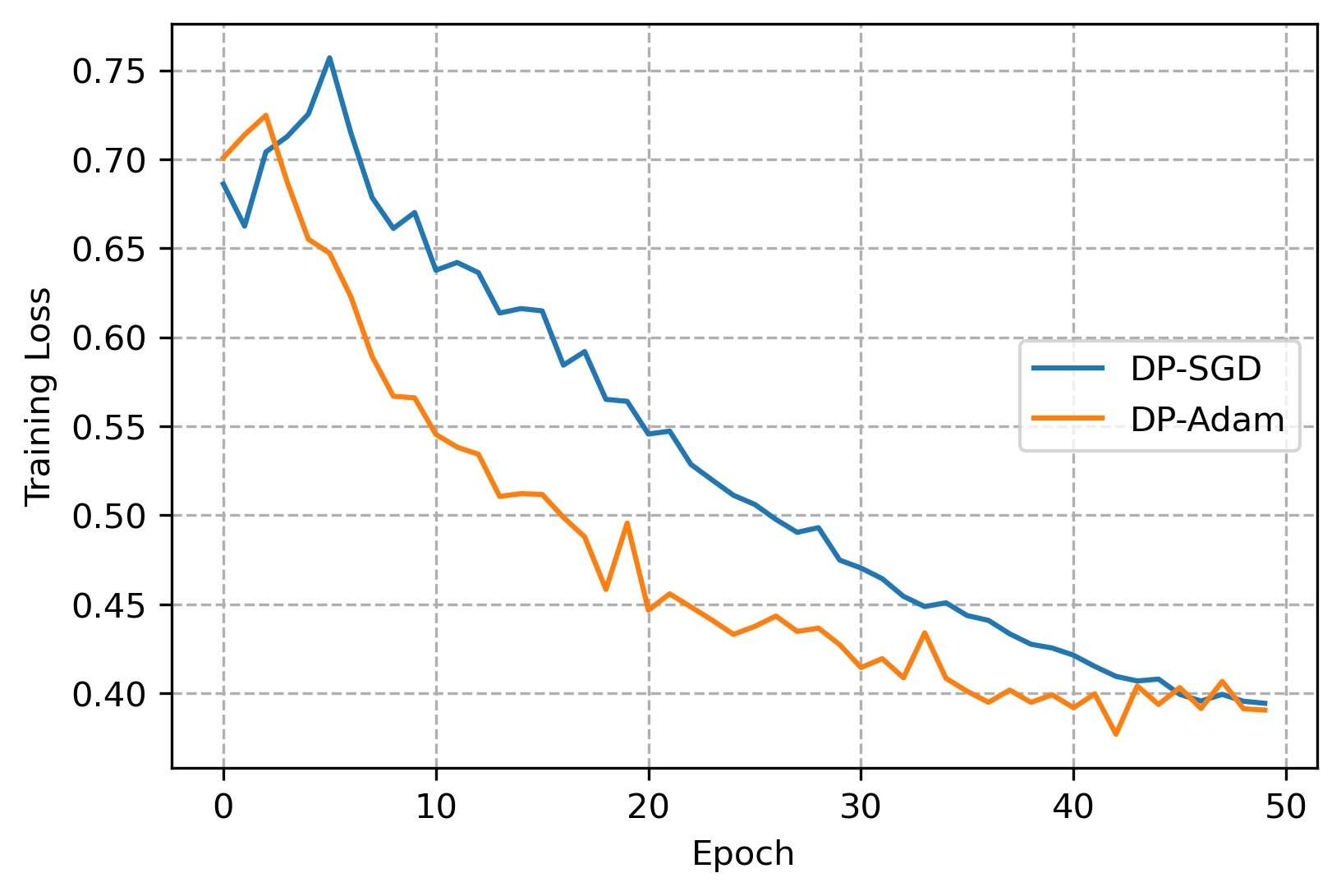}
         \caption[]{Training loss} % <---
         \label{fig:dpadam_vs_dpsgd:train_loss}
     \end{subfigure}
    %  \hfill
     \begin{subfigure}[b]{0.32\textwidth}
         \includegraphics[width=\textwidth]{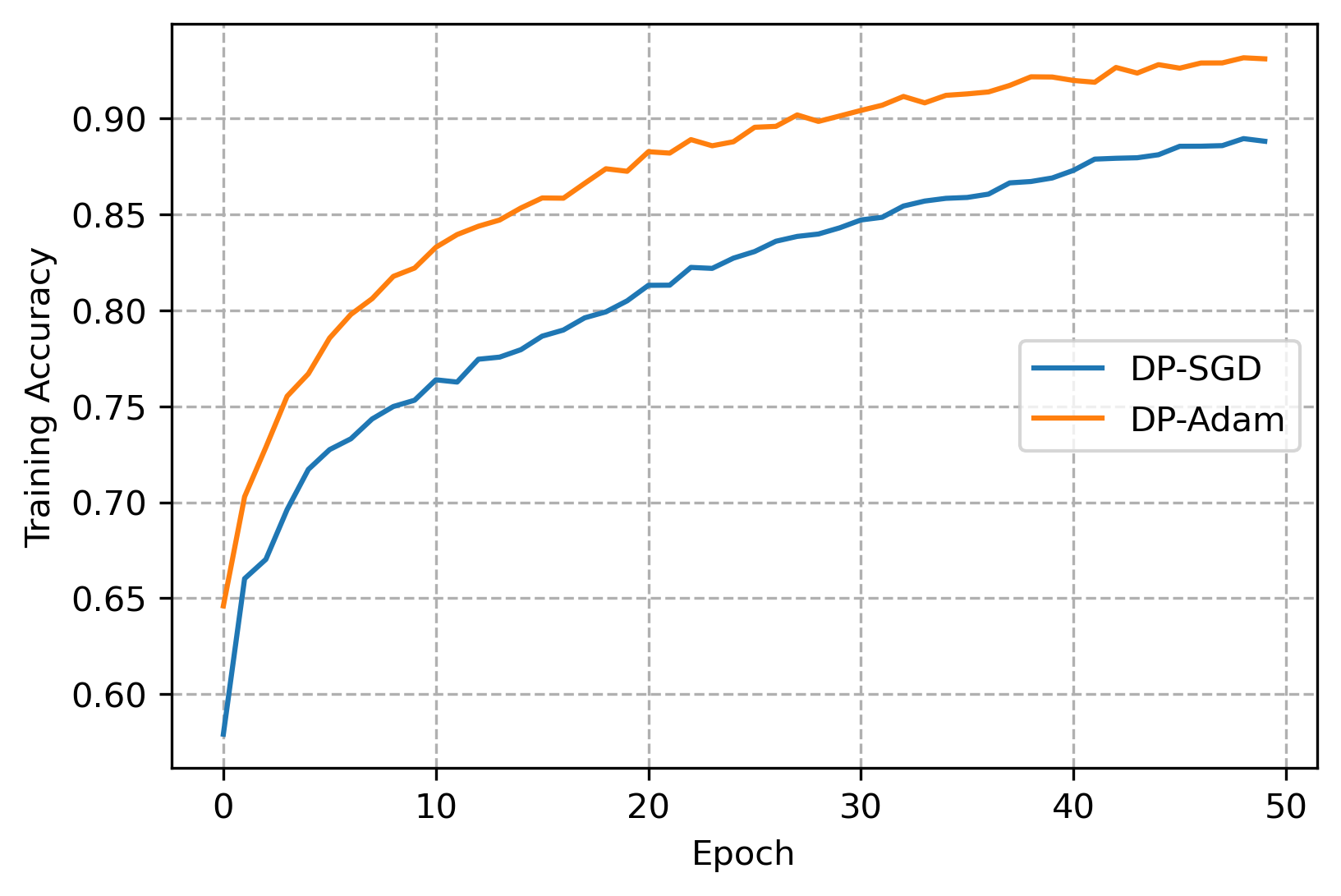}
         \caption[]{Training accuracy} % <---
         \label{fig:dpadam_vs_dpsgd:train_acc}
     \end{subfigure}
     \begin{subfigure}[b]{0.32\textwidth}
         \includegraphics[width=\textwidth]{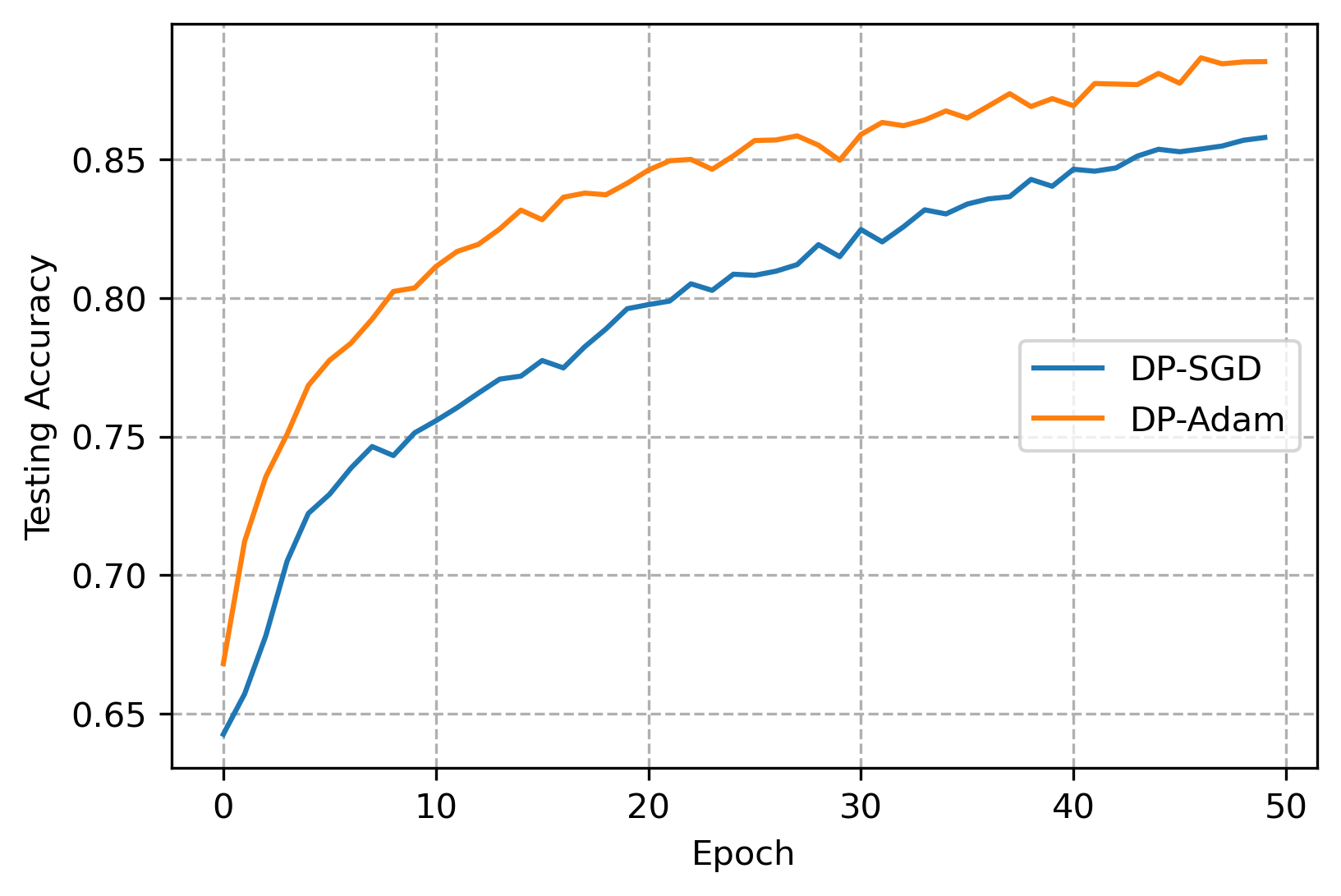}
         \caption[]{Testing accuracy} % <---
         \label{fig:dpadam_vs_dpsgd:test_acc}
     \end{subfigure}
     \caption{Learning curves for DP-SGD and DP-Adam. With ($\epsilon=22.59$, $\delta=10^{-5}$)-differential privacy, we achieve 85.78\% and 88.66\% testing accuracy for DP-SGD and DP-Adam, respectively.} % <---
     \label{fig:dpadam_vs_dpsgd}
\end{figure*}

The training loss, training accuracy, and testing accuracy for the DP optimizers are shown in Figure \ref{fig:dpadam_vs_dpsgd}. We can see that DP-Adam outperforms DP-SGD. Specifically, the highest test accuracy for DP-Adam is 88.66\%, while it is 85.78\% for DP-SGD. We also ran additional experiments with different privacy budgets and found similar results. That is, DP-Adam consistently yields better results than DP-SGD. As such, for all the remaining experiments in this paper, we will use DP-Adam unless otherwise stated. 

\paragraph{Effect of the parameters}

\begin{figure}[htb]
\centering
%\captionsetup[subfigure]{font=small} if you like to change caption style
     \begin{subfigure}[b]{0.45\textwidth}
         \includegraphics[width=\textwidth]{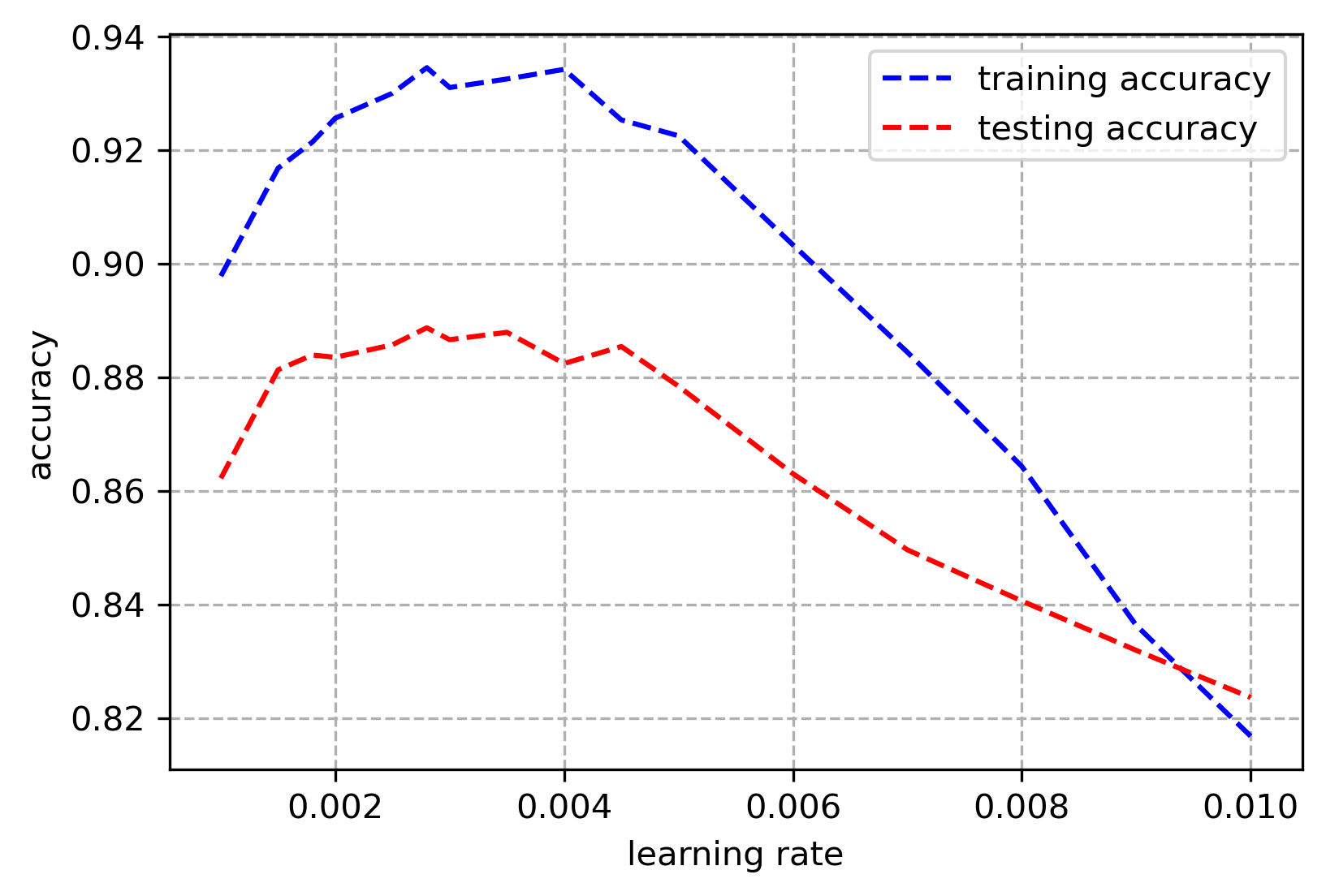}
         \caption[]{Learning rate} % <---
        \label{fig:DP_CT:impact_lr_norm_act:lr}
     \end{subfigure}
    %  \hfill
     \begin{subfigure}[b]{0.45\textwidth}
         \includegraphics[width=\textwidth]{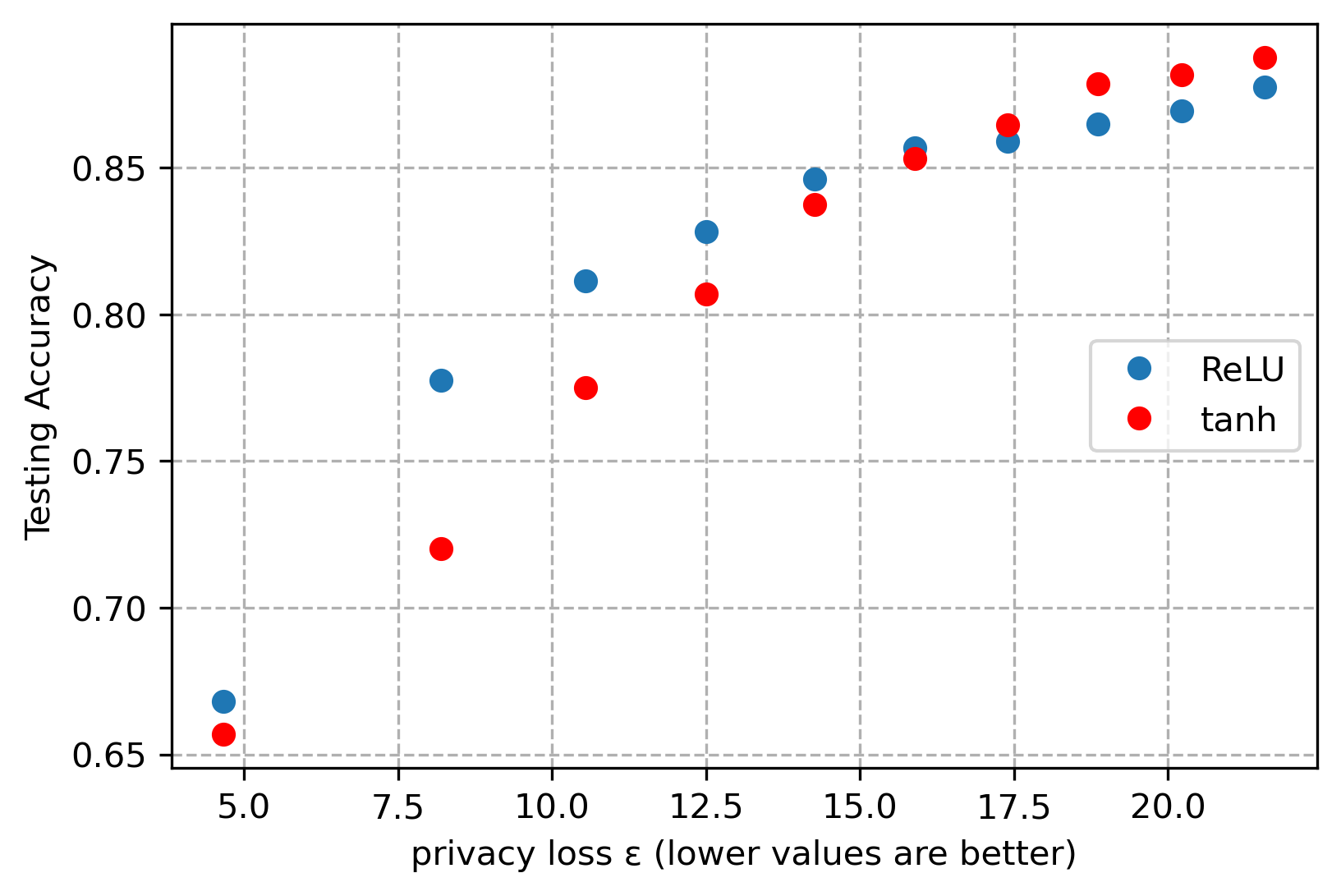}
         \caption[]{Activation function} % <---
        \label{fig:DP_CT:impact_lr_norm_act:act}
     \end{subfigure}
     \begin{subfigure}[b]{0.45\textwidth}
         \includegraphics[width=\textwidth]{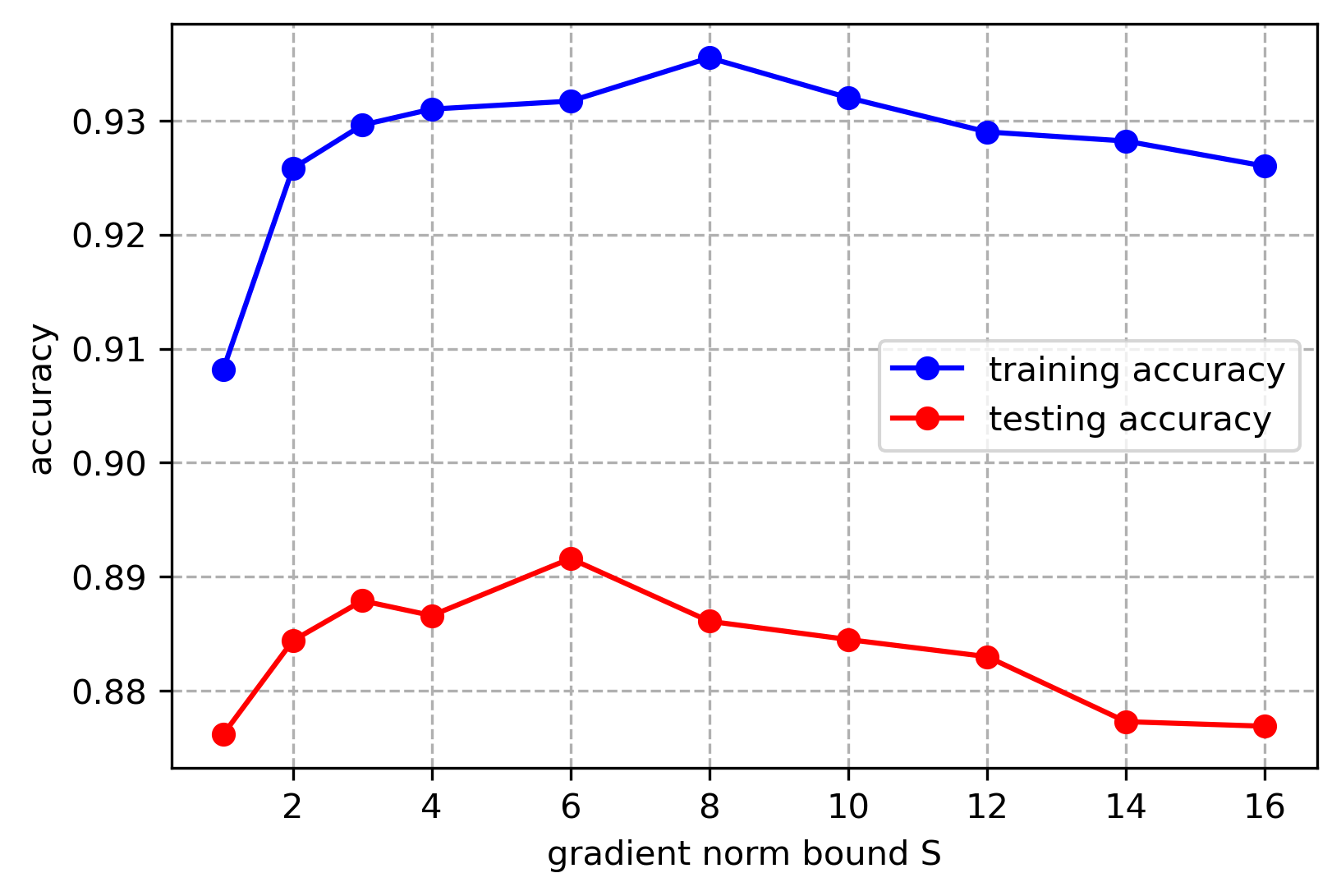}
         \caption[]{Gradient norm bound S} % <---
        \label{fig:DP_CT:impact_lr_norm_act:norm}
     \end{subfigure}
     \begin{subfigure}[b]{0.45\textwidth}
         \includegraphics[width=\textwidth]{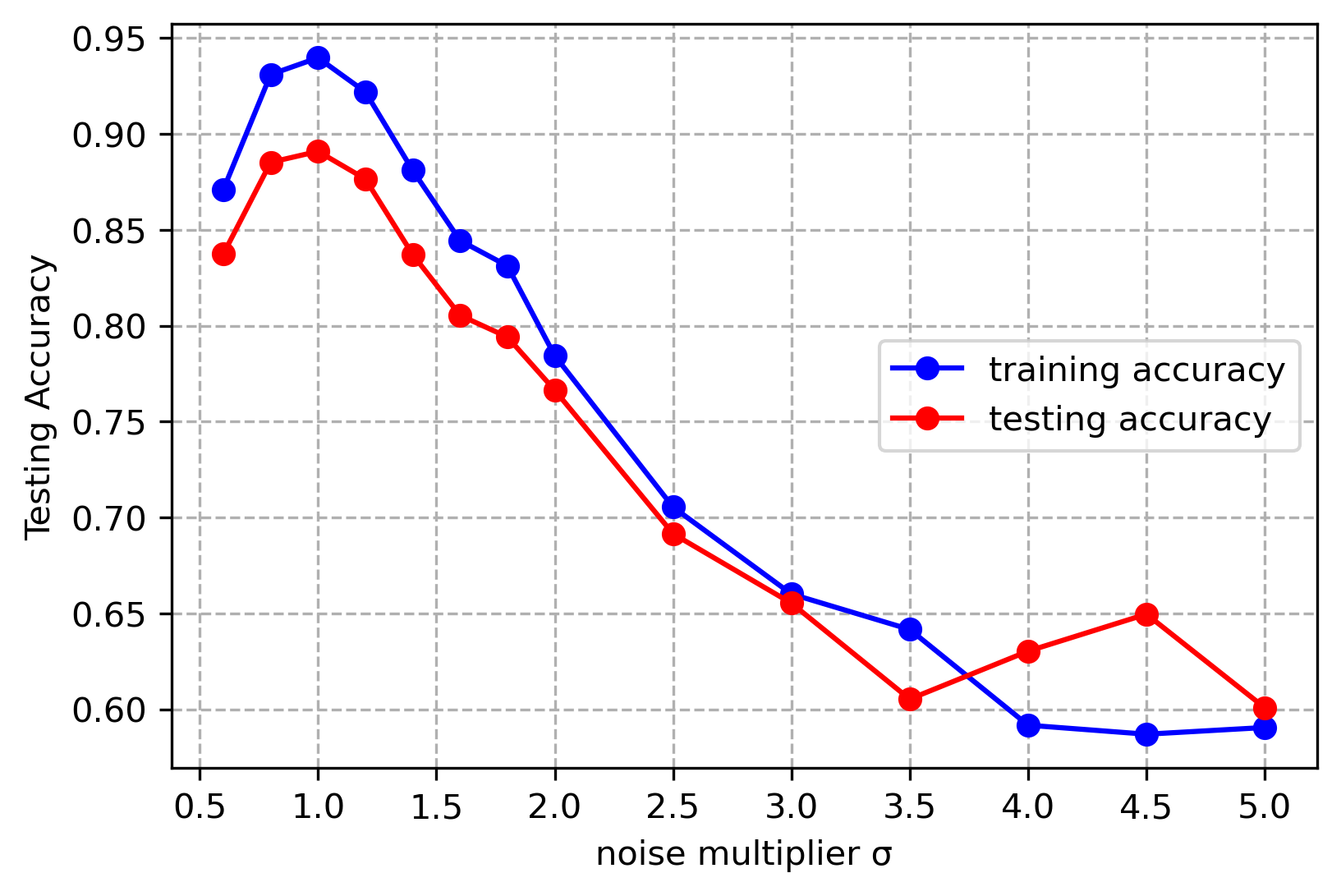}
         \caption[]{Noise multiplier $\sigma$} % <---
        \label{fig:DP_CT:impact_lr_norm_act:sigma}
     \end{subfigure}
     \caption{The effect of various parameters on differentially private training. We vary one parameter while keeping the others constant at reference values. The ($\epsilon$, $\delta$)-DP guarantee is fixed at (22.59, $10^{-5}$) for all the curves.} % <---
     \label{fig:DP_CT:impact_lr_norm_act}
\end{figure}

\paragraph{Learning rate} The learning rate is one of the important hyperparameters that must be fine-tuned to obtain optimal accuracy for differentially private training. The best learning rate for non-private training may not yield optimal results when training a model with DP. We search for the learning rate in the range of [0.001, 0.01]. Figure \ref{fig:DP_CT:impact_lr_norm_act:lr} depicts the accuracy for different learning rates. The test accuracy peaks at the learning rate of 0.003 and decreases for further larger values. Note that we obtained the highest testing accuracy for non-private training at a learning rate of 0.001.

\begin{table}[t!]
    \centering
    \caption{The impact of batch size on differentially private training. All other parameters are fixed at reference values.}
    \label{tab:impact:batch_size_lr}
    \begin{tabular}{|c|c|c|c|c|}
        \hline
        Batch size & Epochs & Learning Rate & Test Acc. & Run Time (s)  \\
        \hline
        512  & 201 & $9.85\cdot 10^{-4}$ & 88.66\% & 333.49 \\
        2048 & 50  & $3\cdot 10^{-3}$    & 88.20\% & 78.3  \\
        \hline
    \end{tabular}
\end{table}

\begin{table}[t!]
    \centering
    \caption{Impact of $\delta$ on differentially private training. The budget for privacy is set at $\epsilon = 22.59$. All other parameters are fixed at reference values.}
    \label{tab:impact:delta}
    \begin{tabular}{|c|c|c|c|c|c|c|}
        \hline
        $\delta$  & $10^{-7}$ & $10^{-6}$ & $10^{-5}$ & $10^{-4}$ & $10^{-3}$ & $10^{-2}$  \\
        \hline
        Test Acc. (\%) & 87.23 & 87.79 & 88.66 & 89.63 & 90.51 & 91.39 \\
        \hline
    \end{tabular}
\end{table}

\paragraph{Batch size} When training with differential privacy, two conflicting objectives must be balanced while determining the minibatch size. On the one hand, smaller batch sizes enable running more epochs, increasing accuracy. On the other hand, the extra noise has a less significant relative impact for larger batch sizes, as increasing the batch size could improve the overall noise-to-signal ratio.

Table \ref{tab:impact:batch_size_lr} compares a model trained for 201 epochs on a batch size of 512 to a model trained for 50 epochs on a batch size of 2048. The total privacy budget for training both models is fixed in both cases ($\epsilon = 22.59$). For both of those batch sizes, we perform a hyperparameter search to fine-tune the choice of the learning rate. The test accuracy obtained with small and large batch sizes is then compared. This experiment shows that training for a small number of epochs at a large batch size can be just as effective as training for a large number of epochs at a small batch size, and the cost in performance is negligible. Furthermore, training a large batch size with a relatively large learning rate is over $4\times$ faster.

\paragraph{Activation function} Gradients computed during optimization are clipped to limit the sensitivity of learning to training examples while training a model with DP. Some of the signal will be lost if these gradients take on large values while gradients are clipped. Preventing the model's activations from exploding is one way to control the magnitude. 
As such, some research \cite{papernot2019making} contends that replacing the unbounded ReLUs with a bounded activation function will prevent activations from exploding and keep the magnitude of gradients to a more reasonable value. This also suggests that the clipping operations applied by DP-Adam (or DP-SGD for that matter) will discard less signal from gradient updates, leading to higher performance at test time.

We train two MLP models with identical architecture, with the exception that the first model uses the ReLU while the second model uses Tanh as activation for its hidden layers. Both models are trained with identical parameters (i.e., the same values for learning rate, batch size, clipping norm $S$, and noise multiplier $\sigma$). We plot the testing accuracy as a function of the privacy loss epsilon in Figure \ref{fig:DP_CT:impact_lr_norm_act:act}. For the fixed privacy budget of $\epsilon=22.59$, the test accuracy of the ReLU model is 88.52\% compared to 89.07\% for the Tanh model. We conducted additional experiments for other privacy budgets and observed that the accuracy of the models for this dataset is not significantly impacted by the choice of activation function.

% The tanh activation function results in only a marginal performance gain for our dataset. 

\paragraph{Clipping bound} One of the fundamental operations used by the differentially private optimizer is clipping each per-example gradient to a maximum fixed $L_2$ norm of $S$. If the gradient norm bound $S$ is too small, the average clipped gradient may point in a direction that is considerably different from the true gradient. On the other hand, since we add noise based on $\sigma S$, increasing the value of $S$ causes us to increase the noise in the gradients (and thus the parameters).

Figure \ref{fig:DP_CT:impact_lr_norm_act:norm} shows the training and testing accuracy for different clipping norms. The testing accuracy peaks at $S=6$ and degrades afterwards. According to \cite{abadi2016deep}, one good way to select a value for $S$ is by taking the median of the norms of the unclipped gradients over the course of training.

\paragraph{Noise multiplier $\sigma$} With more noise, the per-step privacy loss is proportionally smaller, allowing us to run more epochs within a given cumulative privacy budget. We observed that noise multiplier $\sigma$ has a significant impact on model accuracy. Figure \ref{fig:DP_CT:impact_lr_norm_act:sigma} depicts the training and testing accuracy for $sigma$ in the range [0.6, 5.0]. The best test accuracy, 89.09\%, is obtained with $\sigma = 1.0$. It can be seen that when the value of $\sigma$ becomes too large, the learning becomes increasingly difficult, leading to a significant decline in test accuracy.

\paragraph{Impact of $\delta$} When choosing $\delta$, a general rule of thumb is to set it to a value that is less than $1/N$, where N is the size of the training dataset. Because our training dataset contains slightly more than 30,000 samples, we choose $10^{-5}$ as a reference value for $\delta$. We investigate the effect of $\delta$ on the model's accuracy by varying the $\delta$ between $10^{-7}$ and $10^{-2}$ with the privacy budget $\epsilon$ fixed at 22.59. The results of this experiment are shown in Table \ref{tab:impact:delta}. As can be seen, the testing accuracy moderately increases when we increase the $\delta$. However, according to the mathematical definition of DP, a lower $\delta$ value ensures more privacy, and thus increasing $\delta$ degrades privacy.

\paragraph{The trade-off between privacy and accuracy} We study the impact of privacy on model accuracy by training models with different privacy budgets. To achieve stronger privacy (i.e., a lower value of $\epsilon$), we will need to use a high noise multiplier $\sigma$ while training. We experiment with privacy budget $\epsilon \in \{2.5, 4.0, 8.0, 12.0\}$. We set the batch size to 2048, the learning rate to 0.003, the clipping bound $S$ to 4, and $\delta$ to $10^{-5}$ for all the experiments. For each privacy budget $\epsilon$, we train up to the point where our pre-defined privacy budget is violated. For privacy budget $\epsilon$ values of 2.5, 4.0, 8.0, and 12.0, respectively, we choose the noise multiplier $\sigma$ values of 3, 2.3, 1.6, and 1.4. 

\begin{figure}[!t]
\centering
%\captionsetup[subfigure]{font=small} if you like to change caption style
     \begin{subfigure}[b]{0.45\textwidth}
         \includegraphics[width=\textwidth]{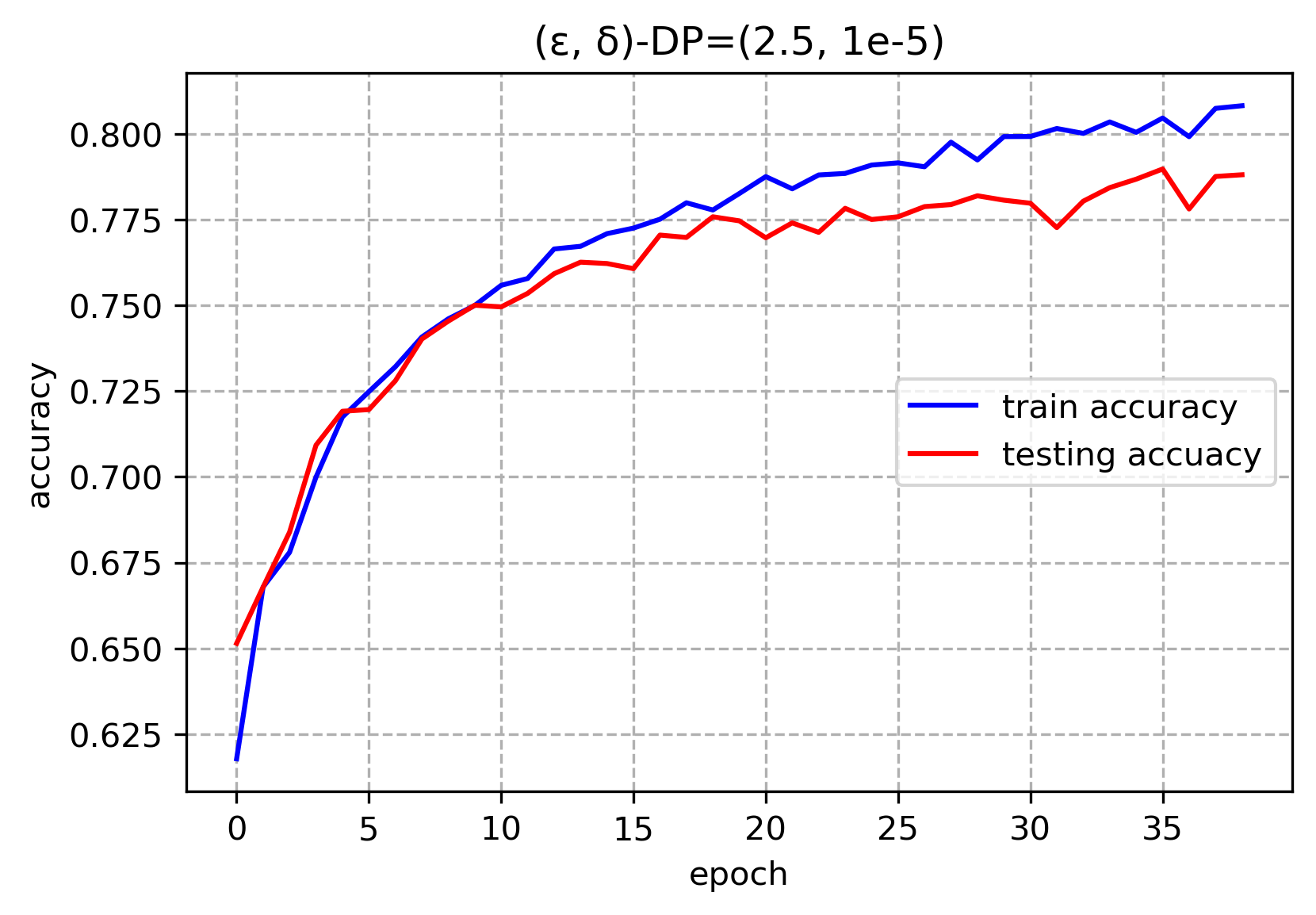}
         \caption[]{$\epsilon=2.5$} % <---
         \label{fig:DP_CT:priv_vs_acc:high}
     \end{subfigure}
    %  \hfill
     \begin{subfigure}[b]{0.45\textwidth}
         \includegraphics[width=\textwidth]{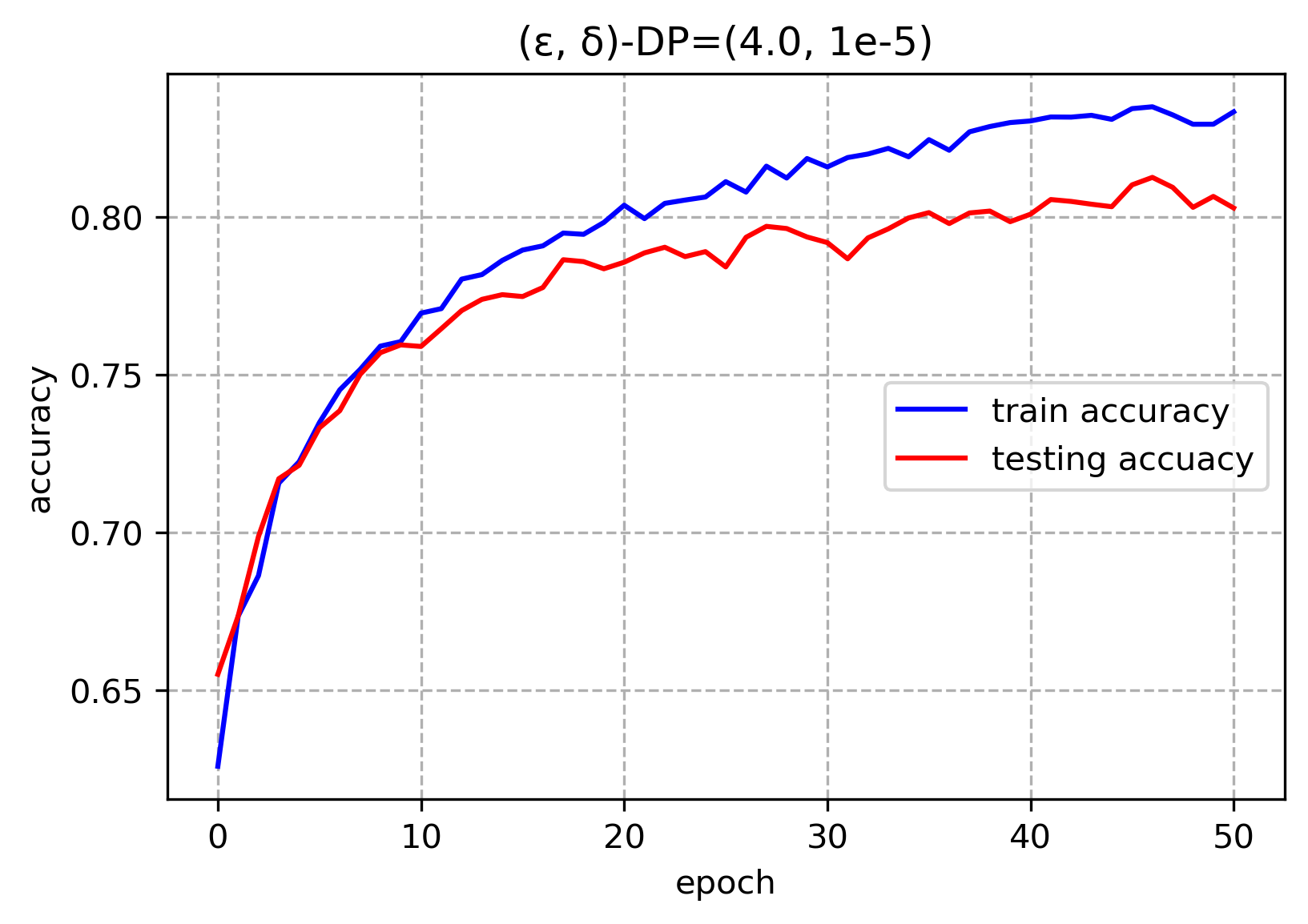}
         \caption[]{$\epsilon=4.0$} % <---
          \label{fig:DP_CT:priv_vs_acc:med}
     \end{subfigure}

     \vskip\baselineskip
     \begin{subfigure}[b]{0.45\textwidth}
         \includegraphics[width=\textwidth]{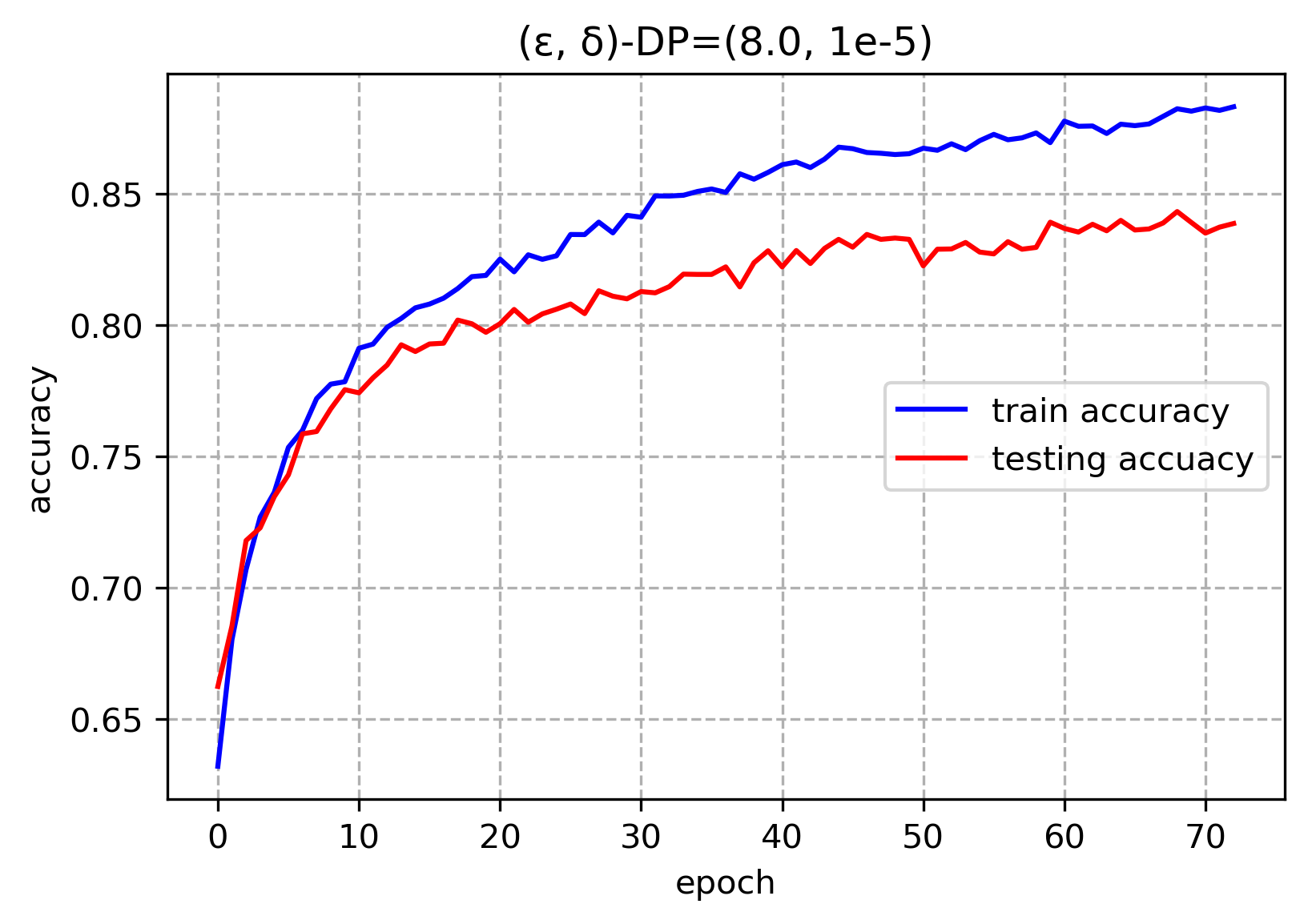}
         \caption[]{$\epsilon=8.0$} % <---
          \label{fig:DP_CT:priv_vs_acc:lower_med}
     \end{subfigure}
    %  \hfill
     \begin{subfigure}[b]{0.45\textwidth}
         \centering
         \includegraphics[width=\textwidth]{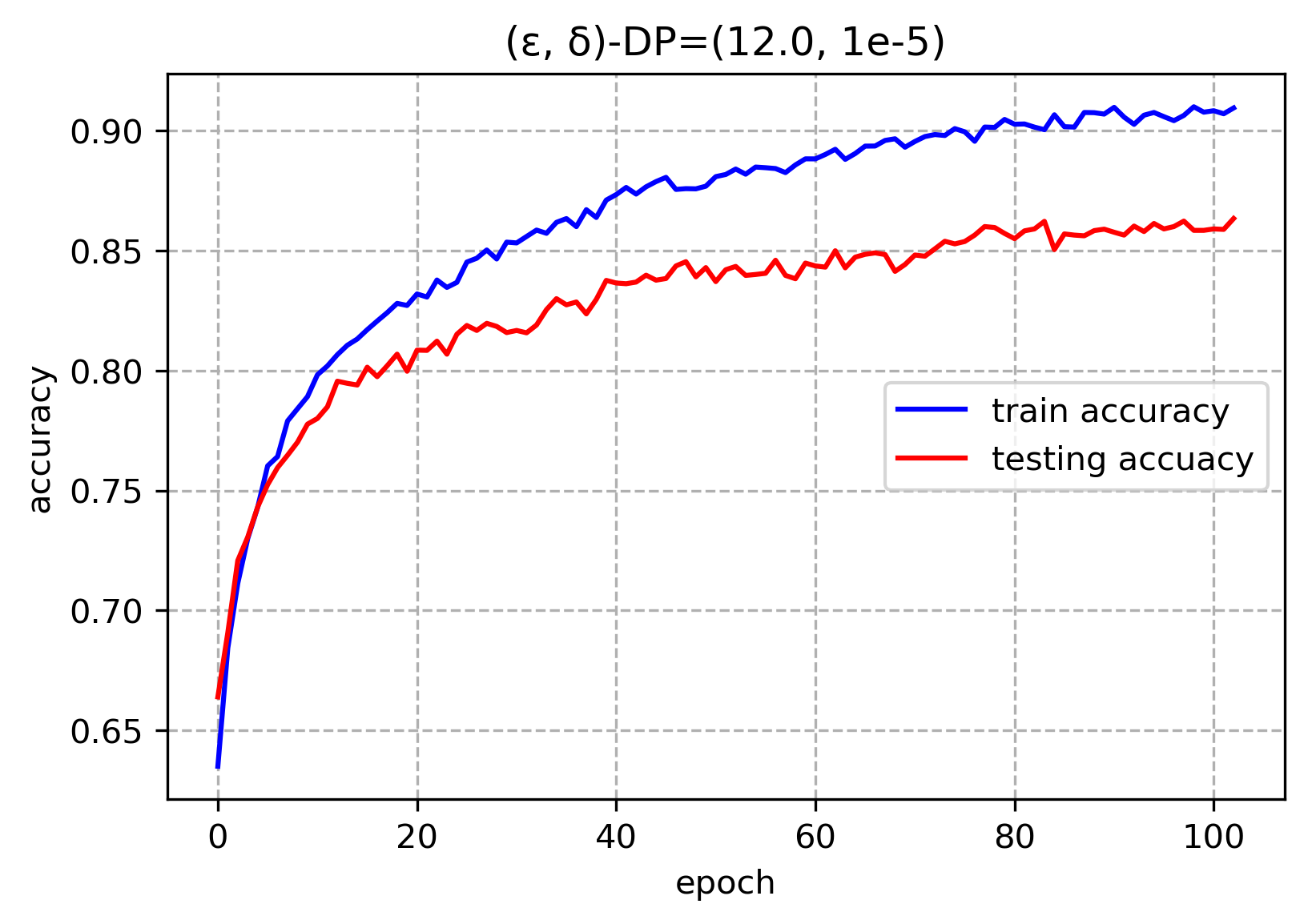}
         \caption[]{$\epsilon=12.0$} % <---
         \label{fig:DP_CT:priv_vs_acc:low}
     \end{subfigure}
        \caption{Results on accuracy for different DP budgets ($\epsilon$, $\delta=10^{-5}$). We achieve a testing accuracy of 78.81\%, 80.28\%, 83.87\%, and 86.33\%, with $\epsilon$ being 2.5, 4.0, 8.0, and 12.0, respectively.} % <---
         \label{fig:DP_CT:priv_vs_acc}
\end{figure}

Figure \ref{fig:DP_CT:priv_vs_acc} visualizes the learning curves for the four privacy budgets. We can see that for smaller noise multiplier $\sigma$ values, we can obtain reasonably high test accuracy. To achieve high privacy or a low value of $\epsilon$, we need to use a larger $\sigma$. However, this negatively impacts the accuracy. As such, while a model with differential privacy, we need to tweak different parameters to find the right balance between privacy and accuracy. We achieve 78.81\%, 80.28\%, 83.87\%, and 86.33\% test accuracy with $\epsilon$ being 2.5, 4.0, 8.0, and 12.0, respectively.

\subsection{Non-private federated learning (baseline)}
We also study the feasibility of federated learning (FL) using the FedAvg algorithm to learn a shared fNIRS classification model between multiple clients without sharing their local fNIRS datasets. FedAvg coordinates training through a central server that hosts the shared global model $w_t$, where $t$ is the communication round. However, the actual optimization is carried out locally on clients using, for example, SGD. The main hyperparameters of the FedAvg algorithm are the following: the total number of clients $K$, the fraction of clients $C$ to select for training, the local mini-batch size $B$, the number of local epochs $E$, and a learning rate $\eta$. Algorithm \ref{alg:fedavg} shows the overview of the FedAvg algorithm. The algorithm starts by randomly initializing the global model $w_0$. A typical communication round of FedAvg consists of the following steps:

\begin{enumerate}
    \item  The server selects a subset of clients $S_t$, $|S_t|=C\cdot K \ge 1$, each of which downloads the current model $w_t$.
    \item Each client in the subset $S_t$ trains the model for $E$ epochs on their respective datasets with a batch size of $B$ and a learning rate of $\eta$.
    \item Clients upload their updated local models $w_{t+1}^k$ to the server.
    \item The server produces the new global model $w_{t+1}$ by computing a weighted sum of all the local models received.
\end{enumerate}

\begin{figure*}[!t]
\centering
%\captionsetup[subfigure]{font=small} 
     \begin{subfigure}[b]{0.32\textwidth}
         \includegraphics[width=\textwidth]{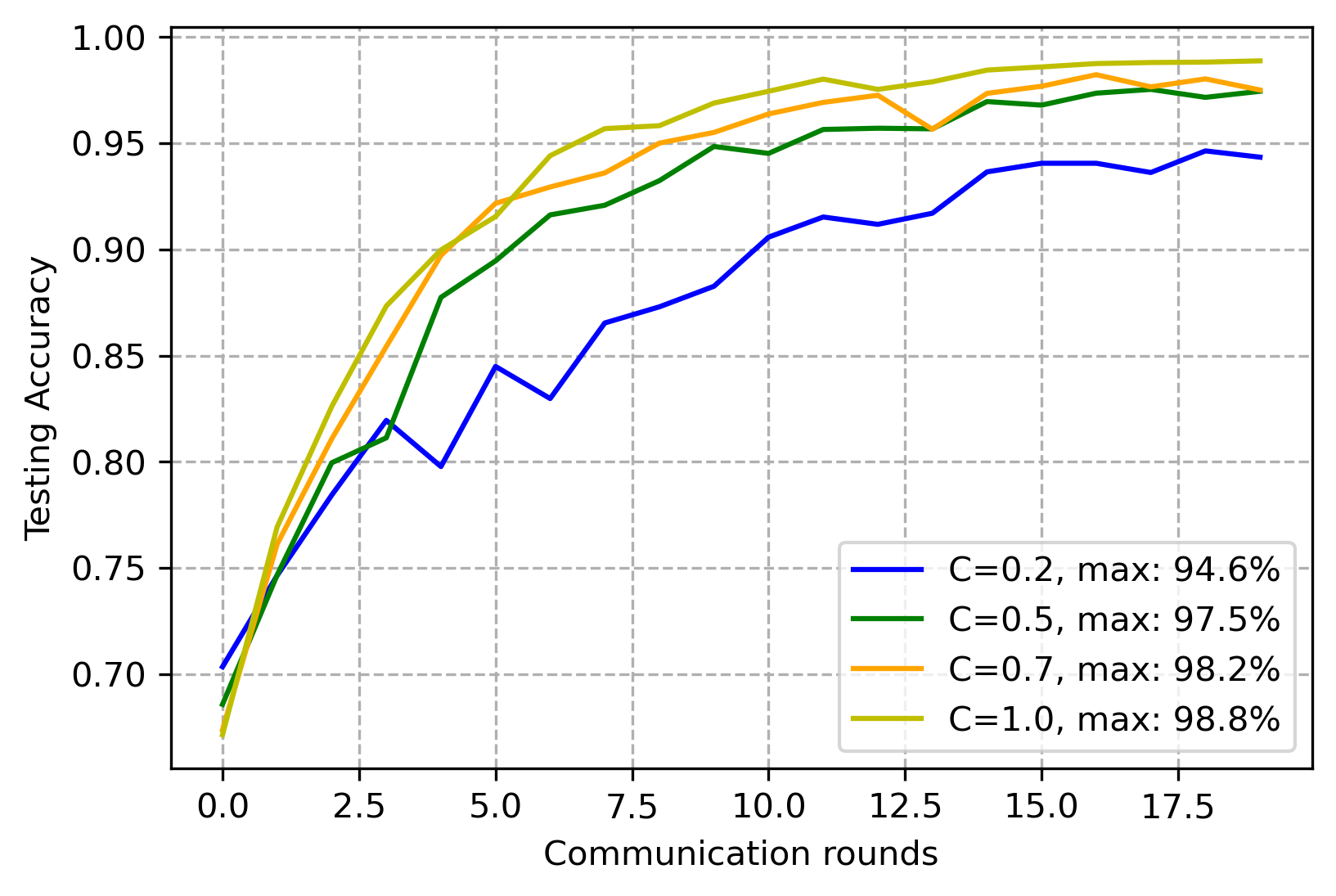}
         \caption[]{$K=5$} % <---
         \label{fig:FL:iid:nb_clients_5}
     \end{subfigure}
    %  \hfill
     \begin{subfigure}[b]{0.32\textwidth}
         \includegraphics[width=\textwidth]{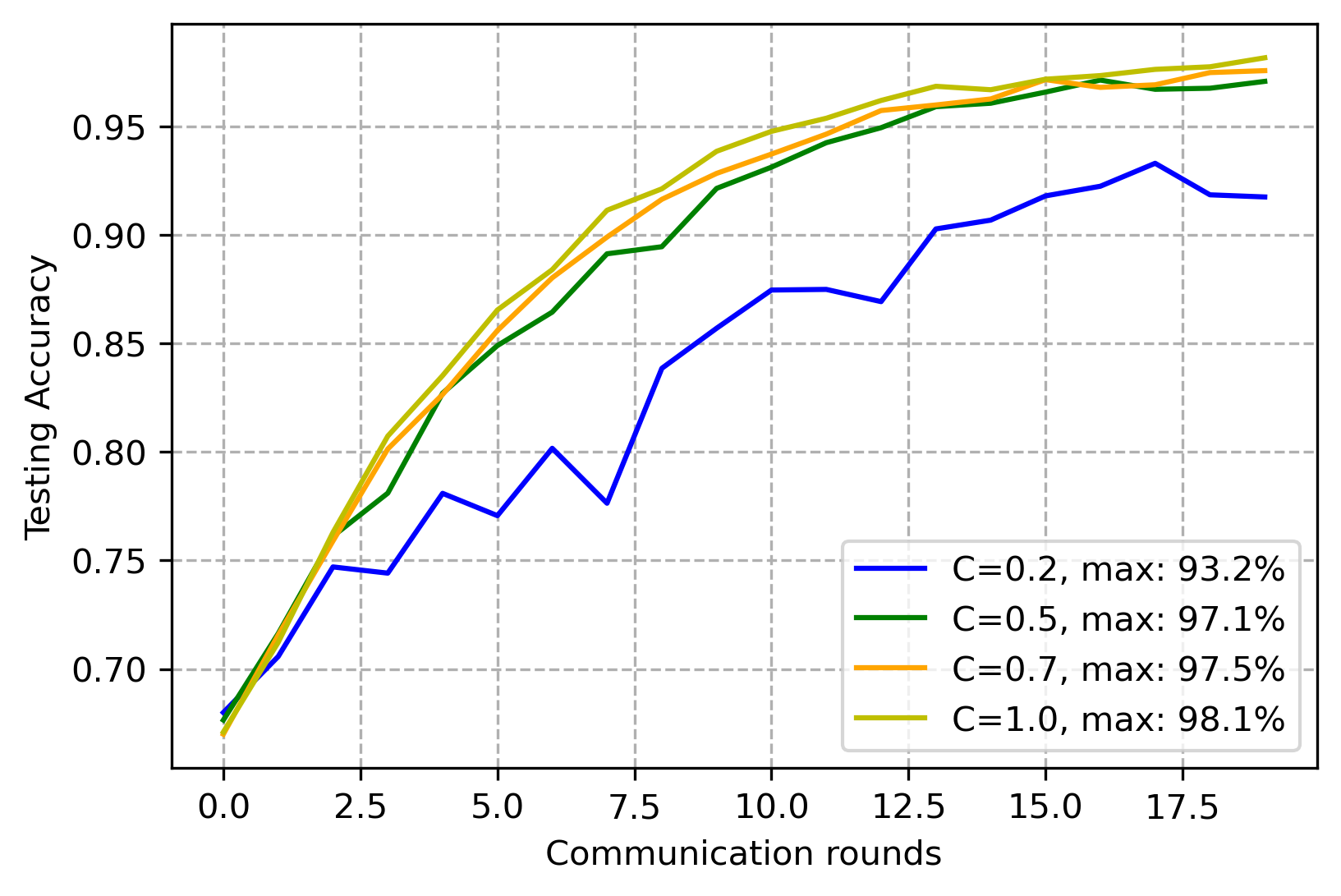}
         \caption[]{$K=10$} % <---
         \label{fig:FL:iid:nb_clients_10}
     \end{subfigure}
     \begin{subfigure}[b]{0.32\textwidth}
         \includegraphics[width=\textwidth]{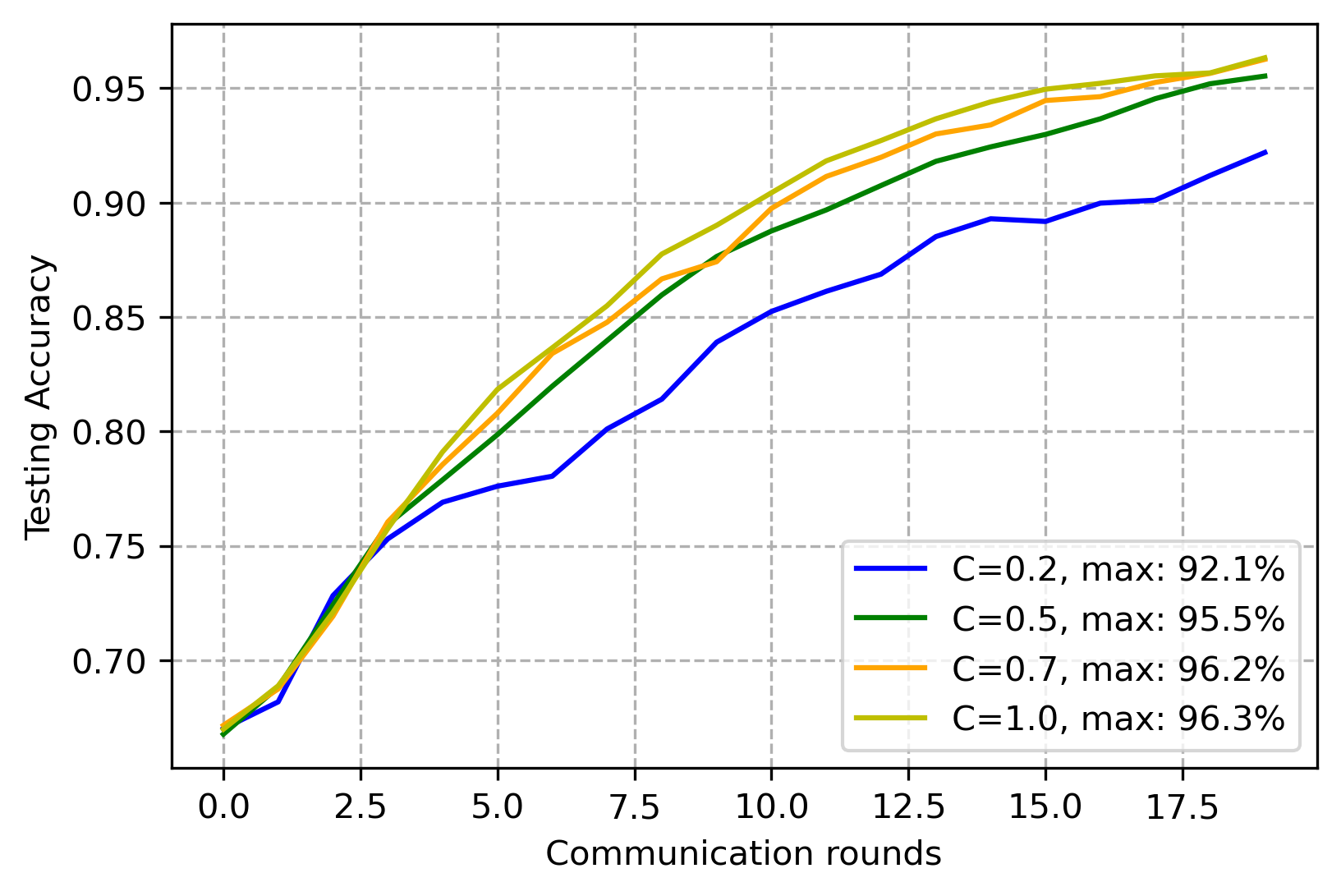}
         \caption[]{$K=20$} % <---
        \label{fig:FL:iid:nb_clients_20}
     \end{subfigure}
     \caption{Testing accuracy of federated learning models in an IID setting with a different number of clients. The number of global communication rounds is set to 20 with $E = 5$, $B = 1024$, and $\eta = 0.003$ for local updates.} % <---
     \label{fig:FL:iid}
\end{figure*}

For this experiment, we used the number of federated clients $K \in \{5,10,20\}$ and the fraction of clients $C \in \{0.2, 0.5, 0.7, 1.0\}$. Moreover, we follow a uniform and independent and identically distributed (IID) setup to distribute the original data set among clients. For each value of $K$, we perform a hyperparameter search to find the optimal values for $B$, $E$, and $\eta$. We found that we can achieve reasonable high test precision for any value of $K$ by running only 20 rounds of FedAvg with $E=5$, $B=1024$, $\eta=0.003$. We also found that updating local models with the Adam optimizer outperforms SGD, which is consistent with our results for centralized training. Figure \ref{fig:FL:iid} visualizes the learning curves for the federated learning model with a different number of clients $K$. Generally, the accuracy of the test increases as $C$ increases. However, we are able to achieve a relatively high test accuracy for $C = 0.5$. Furthermore, accuracy decreases when a larger number of clients $K$ participate in federated learning. With $C = 0.5$, the highest test accuracy for FedAvg with 5, 10, and 20 clients is 98.2\%, 97.5\%, and 96.3\%, respectively. 

\subsection{Differentially private federated learning}
As discussed earlier, while FL attempts to solve some of the security and privacy concerns associated with centralized learning, it still has significant privacy and robustness issues that have been highlighted by earlier research \cite{zhu2019deep, geiping2020inverting, kairouz2021advances}. For example, FL is vulnerable to membership inference attacks \cite{truex2019demystifying, nasr2018comprehensive, shokri2017membership} where the adversary aims to learn if a data point is part of a target's training set. The established framework for defining functions that are not susceptible to adversarial inferences is differential privacy (DP), which allows us to bound the loss of privacy of individual data subjects by adding noise. In the context of FL, two variants of DP can be used: 1) Local DP (LDP) \cite{pihur2018differentially}, in which each participant adds noise before sending updates to the server; and 2) Central DP (CDP) \cite{geyer2017differentially, mcmahan2017learning}, in which the server uses a DP aggregation algorithm.
In this report, we use LDP in FL. The noise addition required for DP in LDP is done locally by each participant. Each participant runs a random perturbation algorithm $M$ and communicates the results to the server. The perturbed result is guaranteed to protect an individual's data according to $\epsilon$.

We implemented LDP in FL. The LDP in FL follows a similar approach to FedAvg. The only difference is that instead of clients running the normal SGD algorithm to update the model, they train the models on their own datasets using differentially private stochastic gradient descent (DP-SGD). This method enables us to use Moments Accountant to keep track of the privacy budget. If a client goes over the pre-defined privacy budget $\epsilon$ set to achieve $\epsilon$-LDP, it will stop updating the model further.

We experiment with the privacy budget $\epsilon \in \{4.0, 8.0, 12.0\}$. The client uses DP-Adam instead of DP-SGD to update the models. We used the same values for the total number of clients $K$ as in non-private FL experiments. However, we set $C$ to 0.7 for $K = 5$ and 0.5 for the other two values of $K$. We set the local batch size $B$ to 256 and the number of local epochs $E$ to 5, the clipping norm $S$ to 4, and $\delta$ to $10^{-5}$ for all experiments. We use the following learning rates: $\eta = 0.003$ for $K = 5$, $\eta = 0.007$ for $K = 10$, and $\eta = 0.01$ for $K = 20$.  Similarly to our non-private FL baselines, we assume that client datasets are uniform and IID.

The results of our experiment are listed in Table \ref{tab:FL_withDP:iid}. We can see that a larger noise multiplier $\sigma$ is required to achieve a lower $\epsilon$ (i.e., stronger privacy) when all other parameters are kept constant. However, increasing the noise $\sigma$ lowers the accuracy. For example, $\epsilon = 12.0$ yields the highest testing accuracy of 78.96\%, and $\epsilon = 4.0$ yields the lowest test accuracy of 73. 02\%. Compared to non-private FL, we can notice that there is a significant discrepancy in the testing accuracy with differentially private FL using LDP. Using DP with $\epsilon = 12.0$, for example, reduces the test by about 19\% compared to the non-private baseline for $K = 5$.

\begin{table}[!tp]
\centering
\caption{Results for differentially private federated learning with LDP. The DP parameter $\delta$ is fixed at $10^{-5}$.}\label{tab:FL_withDP:iid}
\begin{tabular}{ |P{3cm}|P{1cm}|P{1cm}|P{1cm}|P{2cm}|P{3cm}|  }
     \hline
     Privacy Budget & K & C & $\sigma$ & Test Acc. (\%) & Max. Test Acc. (\%) \\
     \hline
     \multirow{3}{4em}{$\epsilon=4.0$} & 5 & 0.7 & 1.7 & 75.59 & 76.03 \\
                                      & 10 & 0.5 & 2.0 & 73.62 & 74.24 \\
                                      & 20 & 0.5 & 2.5 & 73.02 & 73.02 \\
      \hline
      \multirow{3}{4em}{$\epsilon=8.0$} & 5 & 0.7 & 1.2 & 77.66 & 77.98 \\
& 10 & 0.5 & 1.4 & 75.56 & 76.05 \\
& 20 & 0.5 & 1.6 & 74.43 & 74.43\\
      \hline
      \multirow{3}{4em}{$\epsilon=12.0$} & 5 & 0.7 & 0.9 & 78.90 & 78.96\\
& 10 & 0.5 & 1.15 & 76.69 & 77.10\\
& 20 & 0.5 & 1.35 & 75.87 & 75.87\\
      \hline
\end{tabular}
\end{table}

% Since participants train the models on their own datasets using differentially private stochastic gradient descent (DP-SGD), we implement LDP in FL. 

% --- :TODO:--- 
% --> [DONE] describe results for FL with DP.
% --> [DONE] describe exp setting. Add "implim. & eval platform" para/subsec (system, code, etc.)
% --> [DONE] add citations in FL+DP subsec. 
% --> [X] LDP algo? 
% ----> [Inprogress] Reorganize figs, tabs, etc. / fix presentation issues.
% ----> [Inprogress] Polish the overall writing.
% ----> [DONE] Future works sec?? Maybe?
% ----> [DONE] Contribution?

% \section{Low Performance Reason Analysis}

% \section{Improvement Method}

\section{Plan for Next Steps}
% impact of model arch + dimentionality reduction
% FL non-IID
% CDP FL
Deep learning with DP is a relatively new research area. There are still many open problems and opportunities in this area. In this project, we comprehensively studied the impact of various training parameters when learning a deep neural network model with DP. However, we believe that there are several promising directions that merit future research.

First, we used constant values for the noise multiplier and the gradient norm bound in our experiments. Some previous research suggests that dynamically adjusting these two values during training often yields better performance. We will study the effects of doing this in the future.

Second, the architecture of the model plays an important role when learning with privacy. Generally, models with larger parameters perform better than models with a small number of parameters, provided that the size of the training data set is sufficiently large. However, when learning with differential privacy, simpler models with fewer parameters are usually preferred because the norm of the noise vector that DP-SGD (or DP-Adam) must add to the gradient average to maintain privacy increases as the number of parameters increases. In our experiments, we used only a single MLP model with three hidden layers. In the future, we would like to examine other models with various architectures and a different number of parameters to better understand the impact of model architecture on differentially private deep learning.

Third, we assumed that data are distributed in an IID fashion among clients for our federated learning (FL) experiments, which may not be a practical assumption. In a real-world FL setting, clients' training datasets are usually non-IID. As an extension of our current work, we will perform FL in a setting where data are distributed in a non-IID fashion among clients.

Fourth, in this project, we only studied LDP in FL. We found that the accuracy of models trained with LDP is substantially lower than that of their nonprivate counterparts. Utilizing the central DP (CDP) approach for FL might help address this issue and result in better performance. In CDP, the server clips the $L_2$ norm of the updates from the clients, then aggregates the clipped updates, and then adds Gaussian noise to the aggregate. CDP provides DP at the client level, ensuring that the output of the aggregation function is indistinguishable from whether or not a given client is part of the training process, with a probability bounded by $\epsilon$. One of the potential security issues with CDP is that some degree of trust is required in the server. For instance, clients need to trust the server with their model updates and to correctly perform perturbation by adding noise. While there must be some level of trust in the server, this is a considerably weaker assumption than entrusting the server with the data itself. We will investigate the effectiveness of the deferentially private FL using CDP in the future.

\section{Contribution}
In this project, we develop deep learning approaches that preserve privacy for training models on fNIRS datasets. Precisely, to prevent attackers from inferring private fNIRS training data from a learned model, we train models with differential privacy (DP) utilizing differentially private optimization algorithms like DP-SGD and DP-Adam. We extensively investigated how different training parameters affect the accuracy of a model learned with DP. Our study demonstrates that the model accuracy is significantly influenced by learning rate, batch size, noise multiplier $\sigma$, and clipping norm $S$. Our analysis also reveals that the hyperparameters that produce the best results for learning without privacy do not necessarily produce the best results for learning with privacy. As such, performing hyperparameter tuning is very critical to achieving optimal results for privacy-preserving learning. By investigating the trade-off between privacy and utility in private centralized training, we found that we could achieve a testing accuracy of more than 80\% for a privacy budget $\epsilon$ greater than 2.5. We achieved more than 86\% testing precision for a guarantee ($\epsilon$, $\delta$) of (12.0, $10^{-5}$). 

We also examined the effectiveness of federated learning to train a shared fNIRS classification model between multiple clients without sharing their local training datasets. When client training data is IID, we were able to achieve testing accuracy comparable to centralized training. To further improve FL privacy and prevent membership inference attacks, we implemented local DP (LDP) in FL and showed the privacy and accuracy trade-offs. Specifically, we used three values of the privacy budget $\epsilon$ (4, 8 and 8) with different numbers of clients (5, 10, and 20), and our testing accuracy ranged from 73.02\% to 78.90\%. Generally, one would need to sacrifice some privacy to achieve good accuracy.

{\normalsize \bibliographystyle{ieeetr} 
\bibliography{./ref}}

\begin{thebibliography}{10}

\bibitem{dick2021balancing}
E.~Dick, ``Balancing user privacy and innovation in augmented and virtual
  reality,'' tech. rep., Information Technology and Innovation Foundation,
  2021.

\bibitem{disorder}
T.~Ghose, ``Eye tracking could diagnose brain disorders,'' 2012.

\bibitem{hakimi2018stress}
N.~Hakimi and S.~K. Setarehdan, ``Stress assessment by means of heart rate
  derived from functional near-infrared spectroscopy,'' {\em Journal of
  Biomedical Optics}, vol.~23, no.~11, p.~115001, 2018.

\bibitem{hakimi2020proposing}
N.~Hakimi, A.~Jodeiri, M.~Mirbagheri, and S.~K. Setarehdan, ``Proposing a
  convolutional neural network for stress assessment by means of derived heart
  rate from functional near infrared spectroscopy,'' {\em Computers in biology
  and medicine}, vol.~121, p.~103810, 2020.

\bibitem{perdue2014extraction}
K.~L. Perdue, A.~Westerlund, S.~A. McCormick, and C.~A. Nelson~III,
  ``Extraction of heart rate from functional near-infrared spectroscopy in
  infants,'' {\em Journal of biomedical optics}, vol.~19, no.~6, p.~067010,
  2014.

\bibitem{jahani2015attention}
S.~Jahani, N.~H. Berivanlou, A.~Rahimpour, and S.~K. Setarehdan, ``Attention
  level quantification during a modified stroop color word experiment: an fnirs
  based study,'' in {\em 2015 22nd Iranian conference on biomedical engineering
  (ICBME)}, pp.~99--103, IEEE, 2015.

\bibitem{durantin2015characterization}
G.~Durantin, F.~Dehais, and A.~Delorme, ``Characterization of mind wandering
  using fnirs,'' {\em Frontiers in systems neuroscience}, vol.~9, p.~45, 2015.

\bibitem{zhang2016passive}
Z.~Zhang, X.~Jiao, J.~Jiang, J.~Pan, Y.~Cao, H.~Yang, and F.~Xu, ``Passive bci
  based on sustained attention detection: An fnirs study,'' in {\em
  International Conference on Brain Inspired Cognitive Systems}, pp.~220--227,
  Springer, 2016.

\bibitem{murata2015culturally}
A.~Murata, J.~Park, I.~Kovelman, X.~Hu, and S.~Kitayama, ``Culturally
  non-preferred cognitive tasks require compensatory attention: a functional
  near infrared spectroscopy (fnirs) investigation,'' {\em Culture and Brain},
  vol.~3, no.~1, pp.~53--67, 2015.

\bibitem{li2018early}
R.~Li, G.~Rui, W.~Chen, S.~Li, P.~E. Schulz, and Y.~Zhang, ``Early detection of
  alzheimer’s disease using non-invasive near-infrared spectroscopy,'' {\em
  Frontiers in aging neuroscience}, vol.~10, p.~366, 2018.

\bibitem{li2019dynamic}
R.~Li, T.~Nguyen, T.~Potter, and Y.~Zhang, ``Dynamic cortical connectivity
  alterations associated with alzheimer's disease: An eeg and fnirs integration
  study,'' {\em NeuroImage: Clinical}, vol.~21, p.~101622, 2019.

\bibitem{cicalese2020eeg}
P.~A. Cicalese, R.~Li, M.~B. Ahmadi, C.~Wang, J.~T. Francis, S.~Selvaraj, P.~E.
  Schulz, and Y.~Zhang, ``An eeg-fnirs hybridization technique in the
  four-class classification of alzheimer’s disease,'' {\em Journal of
  neuroscience methods}, vol.~336, p.~108618, 2020.

\bibitem{rosas2019prediction}
R.~Rosas-Romero, E.~Guevara, K.~Peng, D.~K. Nguyen, F.~Lesage, P.~Pouliot, and
  W.-E. Lima-Saad, ``Prediction of epileptic seizures with convolutional neural
  networks and functional near-infrared spectroscopy signals,'' {\em Computers
  in biology and medicine}, vol.~111, p.~103355, 2019.

\bibitem{irani2007functional}
F.~Irani, S.~M. Platek, S.~Bunce, A.~C. Ruocco, and D.~Chute, ``Functional near
  infrared spectroscopy (fnirs): an emerging neuroimaging technology with
  important applications for the study of brain disorders,'' {\em The Clinical
  Neuropsychologist}, vol.~21, no.~1, pp.~9--37, 2007.

\bibitem{rizki2015determination}
E.~E. Rizki, M.~Uga, I.~Dan, H.~Dan, D.~Tsuzuki, H.~Yokota, K.~Oguro, and
  E.~Watanabe, ``Determination of epileptic focus side in mesial temporal lobe
  epilepsy using long-term noninvasive fnirs/eeg monitoring for presurgical
  evaluation,'' {\em Neurophotonics}, vol.~2, no.~2, p.~025003, 2015.

\bibitem{zhang2021understanding}
C.~Zhang, S.~Bengio, M.~Hardt, B.~Recht, and O.~Vinyals, ``Understanding deep
  learning (still) requires rethinking generalization,'' {\em Communications of
  the ACM}, vol.~64, no.~3, pp.~107--115, 2021.

\bibitem{song2017machine}
C.~Song, T.~Ristenpart, and V.~Shmatikov, ``Machine learning models that
  remember too much,'' in {\em Proceedings of the 2017 ACM SIGSAC Conference on
  computer and communications security}, pp.~587--601, 2017.

\bibitem{truex2019demystifying}
S.~Truex, L.~Liu, M.~E. Gursoy, L.~Yu, and W.~Wei, ``Demystifying membership
  inference attacks in machine learning as a service,'' {\em IEEE Transactions
  on Services Computing}, 2019.

\bibitem{nasr2018comprehensive}
M.~Nasr, R.~Shokri, and A.~Houmansadr, ``Comprehensive privacy analysis of deep
  learning,'' in {\em Proceedings of the 2019 IEEE Symposium on Security and
  Privacy (SP)}, pp.~1--15, 2018.

\bibitem{shokri2017membership}
R.~Shokri, M.~Stronati, C.~Song, and V.~Shmatikov, ``Membership inference
  attacks against machine learning models,'' in {\em 2017 IEEE symposium on
  security and privacy (SP)}, pp.~3--18, IEEE, 2017.

\bibitem{fredrikson2015model}
M.~Fredrikson, S.~Jha, and T.~Ristenpart, ``Model inversion attacks that
  exploit confidence information and basic countermeasures,'' in {\em
  Proceedings of the 22nd ACM SIGSAC conference on computer and communications
  security}, pp.~1322--1333, 2015.

\bibitem{tramer2016stealing}
F.~Tram{\`e}r, F.~Zhang, A.~Juels, M.~K. Reiter, and T.~Ristenpart, ``Stealing
  machine learning models via prediction $\{$APIs$\}$,'' in {\em 25th USENIX
  security symposium (USENIX Security 16)}, pp.~601--618, 2016.

\bibitem{dwork2011firm}
C.~Dwork, ``A firm foundation for private data analysis,'' {\em Communications
  of the ACM}, vol.~54, no.~1, pp.~86--95, 2011.

\bibitem{dwork2006calibrating}
C.~Dwork, F.~McSherry, K.~Nissim, and A.~Smith, ``Calibrating noise to
  sensitivity in private data analysis,'' in {\em Theory of cryptography
  conference}, pp.~265--284, Springer, 2006.

\bibitem{dwork2014algorithmic}
C.~Dwork, A.~Roth, {\em et~al.}, ``The algorithmic foundations of differential
  privacy,'' {\em Foundations and Trends{\textregistered} in Theoretical
  Computer Science}, vol.~9, no.~3--4, pp.~211--407, 2014.

\bibitem{mcmahan2017communication}
B.~McMahan, E.~Moore, D.~Ramage, S.~Hampson, and B.~A. y~Arcas,
  ``Communication-efficient learning of deep networks from decentralized
  data,'' in {\em Artificial intelligence and statistics}, pp.~1273--1282,
  PMLR, 2017.

\bibitem{vanhaesebrouck2017decentralized}
P.~Vanhaesebrouck, A.~Bellet, and M.~Tommasi, ``Decentralized collaborative
  learning of personalized models over networks,'' in {\em Artificial
  Intelligence and Statistics}, pp.~509--517, PMLR, 2017.

\bibitem{konevcny2016federated}
J.~Kone{\v{c}}n{\`y}, H.~B. McMahan, F.~X. Yu, P.~Richt{\'a}rik, A.~T. Suresh,
  and D.~Bacon, ``Federated learning: Strategies for improving communication
  efficiency,'' {\em arXiv preprint arXiv:1610.05492}, 2016.

\bibitem{bonawitz2019towards}
K.~Bonawitz, H.~Eichner, W.~Grieskamp, D.~Huba, A.~Ingerman, V.~Ivanov,
  C.~Kiddon, J.~Kone{\v{c}}n{\`y}, S.~Mazzocchi, B.~McMahan, {\em et~al.},
  ``Towards federated learning at scale: System design,'' {\em Proceedings of
  Machine Learning and Systems}, vol.~1, pp.~374--388, 2019.

\bibitem{zhu2019deep}
L.~Zhu, Z.~Liu, and S.~Han, ``Deep leakage from gradients,'' {\em Advances in
  neural information processing systems}, vol.~32, 2019.

\bibitem{geiping2020inverting}
J.~Geiping, H.~Bauermeister, H.~Dr{\"o}ge, and M.~Moeller, ``Inverting
  gradients-how easy is it to break privacy in federated learning?,'' {\em
  Advances in Neural Information Processing Systems}, vol.~33,
  pp.~16937--16947, 2020.

\bibitem{wei2020framework}
W.~Wei, L.~Liu, M.~Loper, K.-H. Chow, M.~E. Gursoy, S.~Truex, and Y.~Wu, ``A
  framework for evaluating client privacy leakages in federated learning,'' in
  {\em European Symposium on Research in Computer Security}, pp.~545--566,
  Springer, 2020.

\bibitem{abadi2016deep}
M.~Abadi, A.~Chu, I.~Goodfellow, H.~B. McMahan, I.~Mironov, K.~Talwar, and
  L.~Zhang, ``Deep learning with differential privacy,'' in {\em Proceedings of
  the 2016 ACM SIGSAC conference on computer and communications security},
  pp.~308--318, 2016.

\bibitem{mcmahan2017learning}
H.~B. McMahan, D.~Ramage, K.~Talwar, and L.~Zhang, ``Learning differentially
  private recurrent language models,'' {\em arXiv preprint arXiv:1710.06963},
  2017.

\bibitem{song2013stochastic}
S.~Song, K.~Chaudhuri, and A.~D. Sarwate, ``Stochastic gradient descent with
  differentially private updates,'' in {\em 2013 IEEE global conference on
  signal and information processing}, pp.~245--248, IEEE, 2013.

\bibitem{sahu2018convergence}
A.~K. Sahu, T.~Li, M.~Sanjabi, M.~Zaheer, A.~Talwalkar, and V.~Smith, ``On the
  convergence of federated optimization in heterogeneous networks,'' {\em arXiv
  preprint arXiv:1812.06127}, vol.~3, p.~3, 2018.

\bibitem{li2020federated}
T.~Li, A.~K. Sahu, M.~Zaheer, M.~Sanjabi, A.~Talwalkar, and V.~Smith,
  ``Federated optimization in heterogeneous networks,'' {\em Proceedings of
  Machine Learning and Systems}, vol.~2, pp.~429--450, 2020.

\bibitem{huangfNIRS2MW2021}
Z.~Huang, L.~Wang, G.~Blaney, C.~Slaughter, D.~McKeon, Z.~Zhou, R.~J.~K. Jacob,
  and M.~C. Hughes, ``The tufts fnirs mental workload dataset \& benchmark for
  brain-computer interfaces that generalize,'' in {\em Proceedings of the
  Neural Information Processing Systems (NeurIPS) Track on Datasets and
  Benchmarks}, 2021.

\bibitem{yousefpour2021opacus}
A.~Yousefpour, I.~Shilov, A.~Sablayrolles, D.~Testuggine, K.~Prasad, M.~Malek,
  J.~Nguyen, S.~Ghosh, A.~Bharadwaj, J.~Zhao, {\em et~al.}, ``Opacus:
  User-friendly differential privacy library in pytorch,'' {\em arXiv preprint
  arXiv:2109.12298}, 2021.

\bibitem{schirrmeister2017deep}
R.~T. Schirrmeister, J.~T. Springenberg, L.~D.~J. Fiederer, M.~Glasstetter,
  K.~Eggensperger, M.~Tangermann, F.~Hutter, W.~Burgard, and T.~Ball, ``Deep
  learning with convolutional neural networks for eeg decoding and
  visualization,'' {\em Human brain mapping}, vol.~38, no.~11, pp.~5391--5420,
  2017.

\bibitem{papernot2019making}
N.~Papernot, S.~Chien, S.~Song, A.~Thakurta, and U.~Erlingsson, ``Making the
  shoe fit: Architectures, initializations, and tuning for learning with
  privacy,'' 2019.

\bibitem{kairouz2021advances}
P.~Kairouz, H.~B. McMahan, B.~Avent, A.~Bellet, M.~Bennis, A.~N. Bhagoji,
  K.~Bonawitz, Z.~Charles, G.~Cormode, R.~Cummings, {\em et~al.}, ``Advances
  and open problems in federated learning,'' {\em Foundations and
  Trends{\textregistered} in Machine Learning}, vol.~14, no.~1--2, pp.~1--210,
  2021.

\bibitem{pihur2018differentially}
V.~Pihur, A.~Korolova, F.~Liu, S.~Sankuratripati, M.~Yung, D.~Huang, and
  R.~Zeng, ``Differentially-private" draw and discard" machine learning,'' {\em
  arXiv preprint arXiv:1807.04369}, 2018.

\bibitem{geyer2017differentially}
R.~C. Geyer, T.~Klein, and M.~Nabi, ``Differentially private federated
  learning: A client level perspective,'' {\em arXiv preprint
  arXiv:1712.07557}, 2017.

\end{thebibliography}
\end{document}